\documentclass[journal,onecolumn]{IEEEtran}

\usepackage{epsfig}
\usepackage{graphicx}
\usepackage{amsmath}
\usepackage{amssymb}
\usepackage{commath}
\usepackage{bbding}
\usepackage{cite}
\usepackage{cleveref}
\usepackage{booktabs,multirow}
\usepackage{times}
\usepackage{epstopdf}

\usepackage{mathtools}

\begin{document}

%--------------------------------------------------------------------------------------------- 
%--------------------------------------------------------------------------------------------- 
%--------------------------------------------------------------------------------------------- 
\title{A Closed-Form Model for Image-Based \\Distant Lighting}
%
%
%--------------------------------------------------------------------------------------------- 
%--------------------------------------------------------------------------------------------- 

\author{Mais~Alnasser\thanks{Mais Alnasser was with the Department of Computer Science, University of Central Florida, Orlando, FL, 32816 USA at the time this project was conducted. (e-mail: nasserm@cs.ucf.edu).} and Hassan~Foroosh\thanks{Hassan Foroosh is with the Department of Computer Science, University of Central Florida, Orlando, FL, 32816 USA (e-mail: foroosh@cs.ucf.edu).}
}

\maketitle

%--------------------------------------------------------------------------------------------- 
%--------------------------------------------------------------------------------------------- 
\begin{abstract}
In this paper, we present a new mathematical foundation for
image-based lighting. Using a simple manipulation of the local
coordinate system, we derive a closed-form solution to the light
integral equation under distant environment illumination. We derive
our solution for different BRDF's such as lambertian and
Phong-like. The method is free of noise, and provides the possibility
of using the full spectrum of frequencies captured by images taken
from the environment. This allows for the color of the rendered object
to be toned according to the color of the light in the
environment. Experimental results also show that one can gain an order
of magnitude or higher in rendering time compared to Monte Carlo
quadrature methods and spherical harmonics.
\end{abstract}


%--------------------------------------------------------------------------------------------- 
%--------------------------------------------------------------------------------------------- 
\section{Introduction}

Image-based rendering (IBR) has been an active area of research in computational imaging and computational photography in the past two decades. It has led to many interesting non-traditional problems in image processing and computer vision, which in turn have benefited from traditional methods such as shape and scene description \cite{Cakmakci_etal_2008,Cakmakci_etal_2008_2,Zhang_etal_2015,Lotfian_Foroosh_2017,Morley_Foroosh2017,Ali-Foroosh2016,Ali-Foroosh2015,Einsele_Foroosh_2015,ali2016character,Cakmakci_etal2008,damkjer2014mesh,Junejo_etal_2013,bhutta2011selective,junejo1dynamic,ashraf2007near,Junejo_etal_2007,Junejo_Foroosh_2008,Sun_etal_2012,junejo2007trajectory,sun2011motion,Ashraf_etal2012,sun2014feature,Junejo_Foroosh2007-1,Junejo_Foroosh2007-2,Junejo_Foroosh2007-3,Junejo_Foroosh2006-1,Junejo_Foroosh2006-2,ashraf2012motion,ashraf2015motion,sun2014should},  scene content modeling \cite{Junejo_etal_2010,Junejo_Foroosh_2010,Junejo_Foroosh_solar2008,Junejo_Foroosh_GPS2008,junejo2006calibrating,junejo2008gps,Tariq_etal_2017,Tariq_etal_2017_2,tariq2013exploiting,tariq2015feature,tariq2014scene}, super-resolution (in particular in 3D) \cite{Hu_etal_IBR2012,Foroosh_2000,Foroosh_Chellappa_1999,Foroosh_etal_1996,Cao_etal_2015,berthod1994reconstruction,shekarforoush19953d,lorette1997super,shekarforoush1998multi,shekarforoush1996super,shekarforoush1995sub,shekarforoush1999conditioning,shekarforoush1998adaptive,berthod1994refining,shekarforoush1998denoising,bhutta2006blind,jain2008super,shekarforoush2000noise,shekarforoush1999super,shekarforoush1998blind}, video content modeling \cite{Shen_Foroosh_2009,Ashraf_etal_2014,Ashraf_etal_2013,Sun_etal_2015,shen2008view,sun2011action,ashraf2014view,shen2008action,shen2008view-2,ashraf2013view,ashraf2010view,boyraz122014action,Shen_Foroosh_FR2008,Shen_Foroosh_pose2008,ashraf2012human},  image alignment \cite{Foroosh_etal_2002,Foroosh_2005,Balci_Foroosh_2006,Balci_Foroosh_2006_2,Alnasser_Foroosh_2008,Atalay_Foroosh_2017,Atalay_Foroosh_2017-2,shekarforoush1996subpixel,foroosh2004sub,shekarforoush1995subpixel,balci2005inferring,balci2005estimating,foroosh2003motion,Balci_Foroosh_phase2005,Foroosh_Balci_2004,foroosh2001closed,shekarforoush2000multifractal,balci2006subpixel,balci2006alignment,foroosh2004adaptive,foroosh2003adaptive}, tracking and object pose estimation \cite{Shu_etal_2016,Milikan_etal_2017,Millikan_etal2015,shekarforoush2000multi,millikan2015initialized}, and camera motion quantification and calibration \cite{Cao_Foroosh_2007,Cao_Foroosh_2006,Cao_etal_2006,Junejo_etal_2011,cao2004camera,cao2004simple,caometrology,junejo2006dissecting,junejo2007robust,cao2006self,foroosh2005self,junejo2006robust,Junejo_Foroosh_calib2008,Junejo_Foroosh_PTZ2008,Junejo_Foroosh_SolCalib2008,Ashraf_Foroosh_2008,Junejo_Foroosh_Givens2008,Lu_Foroosh2006,Balci_Foroosh_metro2005,Cao_Foroosh_calib2004,Cao_Foroosh_calib2004,cao2006camera}, to name a few.

Using images to estimate or model environment light for relighting objects introduced or rendered in a scene is a central problem in this area \cite{Cao_etal_2005,Cao_etal_2009,shen2006video,balci2006real,xiao20063d,moore2008learning,alnasser2006image,Alnasser_Foroosh_rend2006,fu2004expression,balci2006image,xiao2006new,cao2006synthesizing}.
This requires solving the light integral equation (also known as the rendering equation), which plays a crucial role in IBR. One of the oldest and most straightforward approaches for
solving the integral is to approximate the solution using the Monte
Carlo method ~\cite{7050}. However, Monte Carlo is an estimation
method, and unless sufficient light samples are taken, it produces
noisy results. Therefore, a substantial number of samples and
accordingly more time is typically required in order to render a
realistic low-noise image.

The key idea that we propose in this paper is the fact that any
light source can be modeled as an area light source. For instance, a
spot light or a directional linear light can both be modeled as
special cases of an area light source, where the dimensionality has
reduced. Similarly, environment lighting using cubemaps may be
viewed as the limiting case of pointwise varying multiple area light
sources. Therefore, in this paper, we first show how the light
integral can be solved in closed-form for a constant area light
source of rectangular shape. We apply our solution to rendering
lambertian and Phong-like materials. We then extend our solution to
non-constant pointwise varying light sources and apply our solution
to lambertian surfaces. In order to streamline the understanding of
the implementation issues, we also provide a pseudocode for our
algorithm.

Because of the closed-form nature of our solution, no sampling is
required and noise is completely eliminated. On the other hand, the
lack of requirement for sampling reduces the rendering time
significantly, making it dependent only on the complexity of the
object (i.e. the number of triangles used to represent it) for a
constant area light source, and dependent on the required highest
light frequency in the case of pointwise varying environment
lighting. In particular, in the case of low-frequency environment
lighting, we achieve the same level of accuracy as spherical
harmonics with coefficients reduced by an order of magnitude in
\emph{O}(1) complexity.  A very simple preprocessing of the light is
required, to compress it using discrete cosine transform, which also
happens to be the most classical tool for image compression.

%-------------------------------------------------------------------------
\section{Related Work}
Some very interesting closed-form solutions have already been
proposed to solve the light integral. The appeal of closed-form
solutions lies in the fact that they provide complete elimination of
noise.  Furthermore, the availability of a closed-form solution
expedites the rendering process significantly. For estimation
methods such as Monte Carlo \cite{226151}, many samples of the
environment are required to render realistic low-noise images. On
the other hand, several hours might be required to generate one
single image.

Currently, the closed-form solutions proposed in the literature
mostly target specific scenarios. For example, the work done in
\cite{Poulin:1991:SSL} targets linear light sources and provides a
solution to the integral for diffuse and specular materials lit by
such a light. The work done by Arvo \cite{arvo-1994a} provides
analytic solutions to the light integral for polyhedral sources
using the irradiance jacobian.  \cite{stark-bsplines} introduced the
use of B-splines to represent surface radiance in static scenes. A
recent work by Sun et al \cite{Sun_2005_5037} provides a closed-form
solution to the light integral given isotropic point light sources.
Their solution targets the scenarios of fog, mist and haze.

Closed-form solutions were also proposed for special cases of
non-constant lighting such as the work done by \cite{732123}, which
provides a solution for linearly-varying luminaires. The most common
application for non-constant lighting is in environment maps.
\cite{383317} used spherical harmonics to solve the light integral
for diffuse materials lit by environment maps, and achieved
real-time rendering.

The methods used to solve the light integral vary and so do the
scenarios for which the light integral is solved. In addition to
advantages described in the previous section in terms of compression
and speed, one of our contributions is to provide a unified
framework that works for spot light, area, and natural distant
environment lighting. We achieve this by formulating a new
closed-form solution to the lighting integral. Our framework works
for lambertian and Phong-like materials. Mirrored, transmissive and
textured materials can be easily embedded within the framework. We
were able to achieve a constant complexity for lambertian materials
depending only on the resolution of the rendered image, suggesting
real-time if implemented on hardware.

\section{Solving the Light Integral}

\begin{figure}[h]
\centering
\includegraphics[width=10cm]{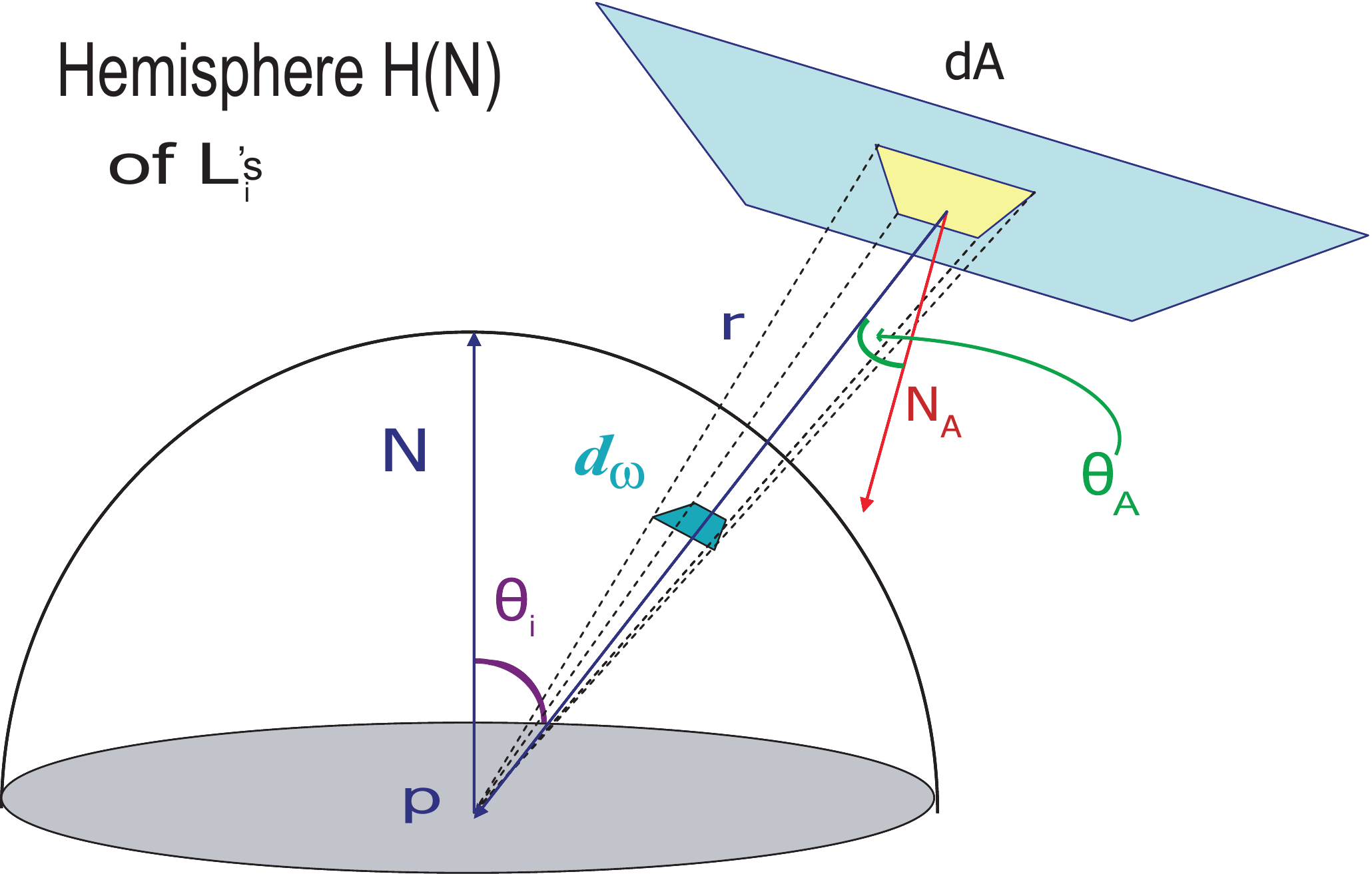}
\caption{Relationship between the incident light and the light
leaving a point on an object surface.} \label{fig:light}
\end{figure}

Figure \ref{fig:light} shows the relationship between the incident
light and the light leaving the surface of an object at a given
point. The following is the the most general form of the light
integral describing this relationship:
\begin{eqnarray}
L_{o}(p,w_{o}) & = & L_e(p,w_o) + \int_{H(N)}f(p,w_o,A_i)\nonumber\\
& & {}L_i(p,A_i)max(0,\cos\theta_i)\frac{\cos\theta_A}{r^2}dA_i
\end{eqnarray}

where $w_o$ is the outgoing direction, $L_e(p,w_o)$ is the emitted
light at point $p$ in the direction $w_o$, $L_o(p,w_o)$ is the
radiance leaving the surface at a point $p$ in the direction $w_o$,
$f(p,w_o,A_i)$ is the Bidirectional Reflectance Distribution
Function (BRDF), $L_i(p,A_i)$ is the incident radiance emitted from
the area patch $A_i$, $\theta_i$ is the angle between the unit
normal $\vec{N}$ at the point $p$ and the vector from $p$ to the
patch $A_i$, $\theta_A$ is the angle between the unit normal $N_A$
of the patch $A_i$ and the light direction, $r$ is the distance
between the point $p$ and the patch $A_i$, $H(\vec{N})$ is the
hemisphere of directions around the normal $\vec{N}$.

In the following sections, we describe our approach to provide
closed-form solutions to this integral equation. We first develop
our approach for lambertian and Phong-like materials in the case of
a rectangular constant area light source, which can naturally reduce
to a point or a linear spot light sources. We then show that the
same approach extends readily to multiple area light sources with
pointwise varying color and intensity, leading thus to a unified
framework that also applies to distant environment lighting using
cubemaps.

\subsection{Constant Area Light Sources}

We start with this case, because its solution shows the underlying
concepts in our approach, from which either specific or more general
cases are derived. We derive this case for lambertian and Phong-like
materials in the following sections.

\subsubsection{Lambertian Materials}\label{ss:lamb}

\begin{figure}[h]
\centering
\includegraphics[width=10cm]{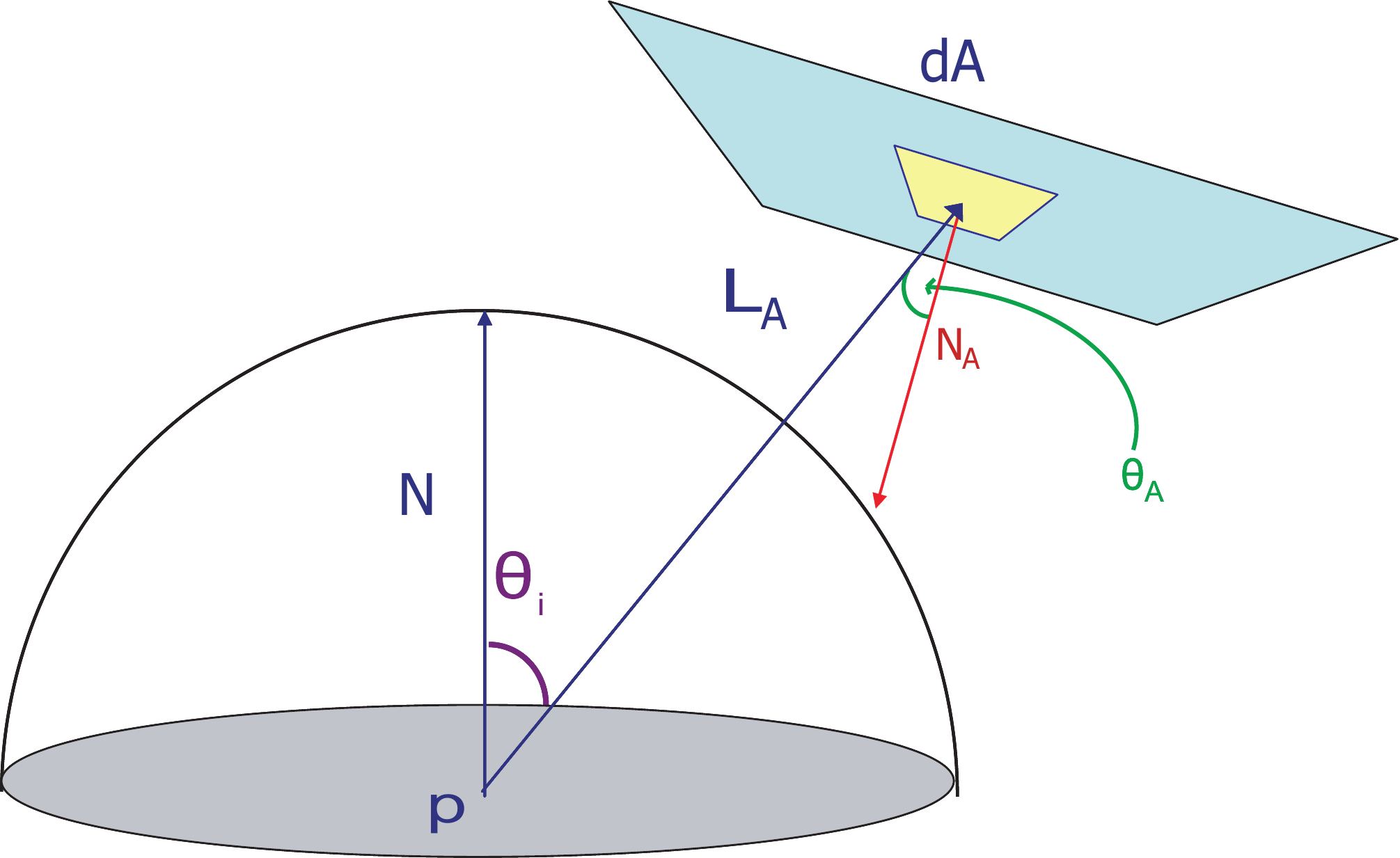}
\caption{Incident and irradiant light on a Lambertian
surface.}\label{fig: Lamb}
\end{figure}

The case for perfectly diffusing Lambertian materials is shown in
Figure \ref{fig: Lamb}. For lambertian materials, the integral
assumes its simplest form:

\begin{equation}
 L_{d}(p) = K_d(p) \;\! I
 \int_{Area}max(0,\cos\theta_i)\frac{\cos\theta_A}{r_A^2}dA
\end{equation}
which can also be written as
\begin{equation}
 L_{d}(p) = k_d \;\! I
 \int_{Area}\frac{max(0,\frac{\vec{L_A}}{r_A}.\vec{N})}{r_A^2}(\frac{-\vec{L_A}}{r_A}.\vec{N_A})dA
\end{equation}

where $k_d=K_d(p)$ is the material albedo as a function of the point
$p$, $I$ is the light intensity, $\vec{N}$ is the unit normal to the
surface at point $p$, $\vec{N_A}$ is the unit normal to the patch
$dA$, $\vec{L_A}$ is the opposite direction of the light emitted by
the patch $dA$, $\theta_i$ is the angle between $\vec{N}$ and
$\vec{L_A}$, $\theta_A$ is the angle between $\vec{N_A}$ and
$-\vec{L_A}$, $r_A$ is the distance between the point $p$ and the
patch $dA$.

Let $d_p=r^2$ be the squared average distance between the area light
source and the point $p$. For typical distant light sources, $r_A$
can be replaced in practice by $r$ with negligible error, in which
case the integral reduces to

\begin{equation}
 L_{d}(p) = \frac{- k_d \;\! I}{d_p^2}
 \int_{Area}max(0,\vec{L_A}.\vec{N})(\vec{L_A}.\vec{N_A})dA
\end{equation}

To simplify the above integral, we transform the endpoints $a_u$,
$b_u$ and $c_u$ of the area light source and its unit normal
$\vec{N_A}_u$ such that the surface point $p$ is at the origin and
the surface unit normal $\vec{N}$ at the point $p$ is along the
Z-axis. We denote the transformed endpoints by $a$, $b$ and $c$, and
the transformed normal by $N_A$.

Let $c_A$ be the center of the patch $dA$ after applying the
transformation, then $L_A$ is the vector $c_A-p$. Since $p$ is
transformed to the point $(0,0,0)$, $L_A$ becomes the vector
$\vec{c_A}=(x_A,y_A,z_A)$, which reduces the integral to

\begin{equation}
 L_{d}(p) = \frac{- k_d \;\! I}{d_p^2}
 \int_{Area}max(0,\vec{c_A}.\vec{N})(\vec{c_A}.\vec{N_A})dA
\end{equation}

Since $\vec{c_A}=(x_A,y_A,z_A)$ and $\vec{N}=(0,0,1)$, we have
$\vec{N}.\vec{c_A}=z_A$, and the integral becomes

\begin{equation}
 L_{d}(p) = \frac{-k_d \;\! I}{d_p^2}
 \int_{Area}max(0,z_A)(\vec{c_A}.\vec{N_A})dA
\end{equation}

The point $c_A$ on the rectangular area light source can be
described in parametric form by $\vec{c_A}=a+u(b-a)+v(c-a)$, where
$0 \leq u,v \leq 1$. The integral now becomes

\begin{eqnarray}
  L_{d}(p) \!\!\!\!\!\!\!\!\!\!\!\!\!&& =
  \frac{-k_d \;\! I}{d_p^2}\int_{u}\int_{v}max(0,a_z+u(b_z-a_z)+v(c_z-a_z)) \nonumber\\
  \!\!\!\!& &\!\!\!\! {} [N_{A_x} N_{A_y} N_{A_z}]\left[
                            \begin{array}{c}
                            a_x+u(b_x-a_x)+v(c_x-a_x) \\
                            a_y+u(b_y-a_y)+v(c_y-a_y) \\
                            a_z+u(b_z-a_z)+v(c_z-a_z) \\
                            \end{array}
                    \right] dv du \nonumber \\
  \!\!\!\!& &\!\!\!\! {}
\end{eqnarray}

\begin{figure}[h]
\centering 
%\vspace*{-1.2cm}
\includegraphics[width=12cm]{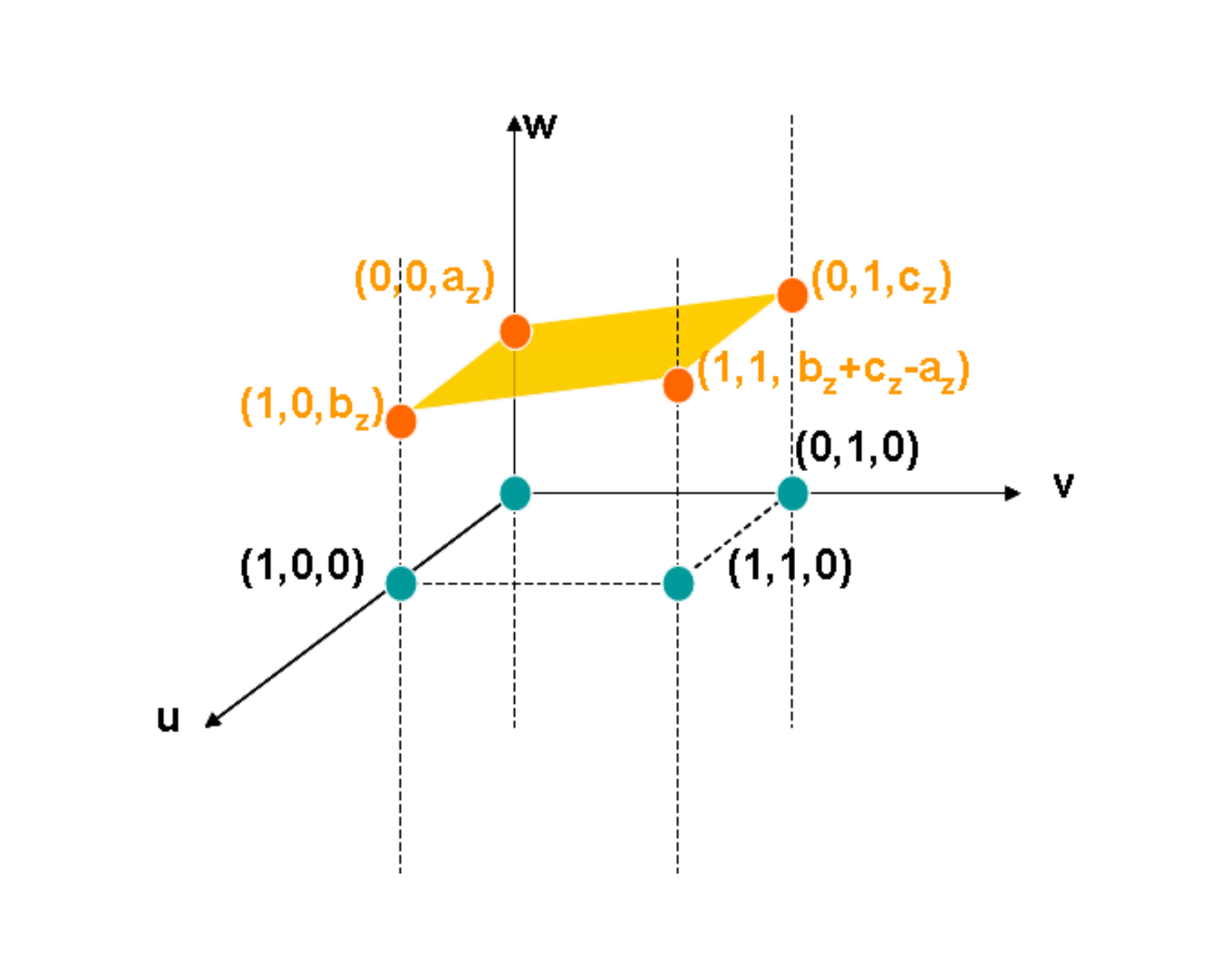}
%\vspace*{-0.5cm} 
\caption{The plane defining the limits of the light
integral.}\label{fig:plane}
\end{figure}

The above formula contains a max term that needs to be eliminated in
order to be able to integrate in closed-form. The key observation
that allows us to do this is the fact that $w=a+u(b-a)+v(c-a)$ is
the equation of a plane as shown in Figure \ref{fig:plane}. As $u$
and $v$ vary in the unit interval, a plane segment is defined, whose
corners are given by
\begin{center}
\begin{tabular}{|c|c|c|}
  \hline
  u & v & w \\
  \hline \hline
  0 & 0 & $a_z$ \\ \hline
  0 & 1 & $c_z$ \\ \hline
  1 & 0 & $b_z$ \\ \hline
  1 & 1 & $b_z+c_z-a_z$ \\
  \hline
\end{tabular}
\end{center}

The max term can be eliminated by identifying the portion of this
plane segment that lies above the plane $w=0$, i.e. by setting the
bounds $u_0\leq u\leq u_1$ and $v_0\leq v\leq v_1$ of the integral
such that only the portion above the plane $w=0$ is taken into
account. This is verified per point on the object being rendered.
There are two trivial cases. The first is when all of the plane
segment is above the plane $w=0$. In this case, the bounds would be
$u_0=v_0=0$, and $u_1=v_1=1$. The second case would be it is all
under the plane $w=0$. In this case the integral evaluates to zero.

Table \ref{tab:bounds} shows the bounds for each of the other cases,
where $u_{line}=\frac{c_z-a_z}{a_z-b_z}v+\frac{a_z}{a_z-b_z}$,
$v_{line}=\frac{b_z-a_z}{a_z-c_z}v+\frac{a_z}{a_z-c_z}$ and
$d_z=b_z+c_z-a_z$. Note that the bounds of the integral are given in
terms of the transformed coordinates of the end points of the area
light source, which are known values. For some cases, the integral
has to be divided into two subintegrals. For those cases the bounds
for the first subintegral are denoted by $u_0$, $u_1$, $v_0$ and
$v_1$ and the bounds for the second subintegral are denoted by
$u_{00}$, $u_{11}$, $v_{00}$ and $v_{11}$.

\begin{table}[t]
\begin{center}
\begin{tabular}{|c|@{}c@{}|@{}c@{}|@{}c@{}|@{}c@{}|@{}c@{}|@{}c@{}|@{}c@{}|@{}c@{}|}
  \hline
                          & $u_0$ & $u_1$ & $v_0$ & $v_1$ & $u_{00}$ & $u_{11}$ & $v_{00}$ & $v_{11}$ \\ \hline \hline
  $a_z\geq0$, $b_z<0$,    &   &   &   &   &   &   &   &   \\
  $c_z<0$, $d_z<0$        & 0 & $u_{line}$ & 0 & $\frac{a_z}{a_z-c_z}$ &   &   &   &   \\ \hline
  $a_z\geq0$, $b_z<0$,    &   &  &   &   &   &   &   &   \\
  $c_z\geq0$, $d_z<0$     & 0 & $u_{line}$ & 0 & 1 &   &   &   &   \\ \hline
  $a_z\geq0$, $b_z<0$,    &   &   &   &   &   &   &   &   \\
  $c_z\geq0$, $d_z\geq0$  & 0 & $u_{line}$ & 0 & $\frac{b_z}{a_z-c_z}$  & 0 & 1 & $\frac{b_z}{a_z-c_z}$ & 1 \\ \hline
  $a_z\geq0$, $b_z\geq0$, &   &   &   &   &   &   &   &   \\
  $c_z<0$, $d_z<0$        & 0 & 1 & 0 & $v_{line}$ &   &  &  &  \\ \hline
  $a_z\geq0$, $b_z\geq0$, &   &   &   &   &   &   &   &   \\
  $c_z<0$, $d_z\geq0$     & 0 & 1 & 0 & $\frac{a_z}{a_z-c_z}$ & $u_{line}$ & 1 & $\frac{a_z}{a_z-c_z}$ & 1 \\ \hline
  $a_z\geq0$, $b_z\geq0$, &   &   &   &   &   &   &   &   \\
  $c_z\geq0$, $d_z<0$     & 0 & 1 & 0 & $\frac{b_z}{a_z-c_z}$ & 0 & $u_{line}$ & $\frac{b_z}{a_z-c_z}$ & 1 \\ \hline
  $a_z<0$, $b_z\geq0$,    &   &   &   &   &   &   &   &   \\
  $c_z\geq0$, $d_z\geq0$  & $u_{line}$ & 1 & 0 & $\frac{a_z}{a_z-c_z}$ & 0 & 1 & $\frac{a_z}{a_z-c_z}$ & 1 \\ \hline
  $a_z<0$, $b_z\geq0$,    &   &   &   &   &   &   &   &   \\
  $c_z<0$, $d_z\geq0$     & $u_{line}$ & 1 & 0 & 1 &  &  &  &  \\ \hline
  $a_z<0$, $b_z\geq0$,    &   &   &   &   &   &   &   &   \\
  $c_z<0$, $d_z<0$        & $u_{line}$ & 1 & 0 & $\frac{b_z}{a_z-c_z}$ &  &  &  &  \\ \hline
  $a_z<0$, $b_z<0$,       &   &   &   &   &   &   &   &   \\
  $c_z\geq0$, $d_z\geq0$  & 0 & 1 & $v_{line}$ & 1 &  &  &  &  \\ \hline
  $a_z<0$, $b_z<0$,       &   &   &   &   &   &   &   &   \\
  $c_z\geq0$, $d_z<0$     & 0 & $u_{line}$ & $\frac{a_z}{a_z-c_z}$ & 1 &  &  &  &  \\ \hline
  $a_z<0$, $b_z<0$,       &   &   &   &   &   &   &   &   \\
  $c_z<0$, $d_z\geq0$     & $u_{line}$ & 1 & $\frac{b_z}{a_z-c_z}$ & 1 &  &  &  &  \\ \hline
  \hline
\end{tabular}
\end{center}
\caption{Table of integral bounds.}\label{tab:bounds}
\end{table}

Once the bounds of the integral are determined as discussed above,
the max term is simply eliminated by the fact that we would be
integrating only for the cases where $w$ is always strictly
positive. Rearranging the terms in the integral then would yield
\begin{eqnarray*}
  L_{d}(p) & = & \frac{- k_d \;\! I}{d_p^2}\int_{u=u_{0}}^{u_1}\int_{v=v_{0}}^{v_1}(l_{00}+l_{01}v+l_{02}v^2+\nonumber\\
  & & {} l_{10}u+l_{11}uv+l_{20}u^2) dv du
\end{eqnarray*}
%\vspace*{-0.8cm}
\begin{eqnarray}
  l_{00} & = & a_z \times (\vec{a}.\vec{N_A}) \nonumber \\
  l_{01} & = & a_z \times ((\vec{c}-\vec{a}).\vec{N_A})+(c_z-a_z)\times (\vec{a}.\vec{N_A}) \nonumber \\
  l_{02} & = & (c_z-a_z) \times ((\vec{c}-\vec{a}).\vec{N_A}) \nonumber \\
  l_{10} & = & a_z \times ((\vec{b}-\vec{a}).\vec{N_A})+(b_z-a_z)\times (\vec{a}.\vec{N_A}) \nonumber \\
  l_{11} & = & (c_z-a_z) \times ((\vec{c}-\vec{a}).\vec{N_A}) + (c_z-a_z) \times ((\vec{b}-\vec{a}).\vec{N_A}) \nonumber \\
  l_{20} & = & (b_z-a_z) \times ((\vec{b}-\vec{a}).\vec{N_A})
\end{eqnarray}
which is simply an integral of a polynomial that can be easily
evaluated in closed-form. The resultant solution can then be
embedded in the rendering code achieving a complexity of
\emph{O}(1).

\subsubsection{Phong-Like Materials}
For Phong-like materials the light integral becomes as follows:
\begin{equation}
 L_{s}(p) = - k_s \;\! I
 \int_{Area}\frac{max(0,\frac{\vec{L_A}}{r_A}.\vec{R})^{sh}}{r_A^2}(\frac{\vec{L_A}}{r_A}.\vec{N_A})dA
\end{equation}

Where $k_s$ is the albedo, $sh$ is the shininess of the material,
and $\vec{R}$ is the reflection of the viewing vector at a point $p$
on the surface.

Substituting $r_A^2$ with $d_p$, the integral becomes
\begin{equation}
 L_{s}(p) = \frac{- k_s \;\! I}{d_p^{(sh+3)/2}}
 \int_{Area}max(0,(\vec{L_A}.\vec{R})^{sh})(\vec{L_A}.\vec{N_A})dA
\end{equation}

For Phong-like materials, we similarly transform the endpoints
$a_u$, $b_u$ and $c_u$ of the area light source and its unit normal
$\vec{N_{A_u}}$ such that the surface point $p$ is at the origin and
the reflection vector $\vec{R}$ at the point $p$ is along the
z-axis.

To simplify the integral, we proceed as we did with the Lambertian
materials. The integral becomes
\begin{equation}
 L_{s}(p) = \frac{- k_s \;\! I}{d_p^{(sh+3)/2}}
 \int_{Area}max(0,(z_A)^{sh})(\vec{c_A}.\vec{N_A})dA
\end{equation}

Again, to eliminate the max term, we divide the integral into the
same cases mentioned above. The integral now becomes
\begin{eqnarray}
  L_{s}(p) \!\!\!\!\!& = &\!\!\!\!\! \frac{- k_s \;\! I}{d_p^{(sh+3)/2}}\!\!\int_{u=u_{0}}^{u_1}\!\!\int_{v=v_{0}}^{v_1}(a_z+u(b_z-a_z)+v(c_z-a_z))^{sh} \nonumber\\
  \!\!\!\!\!\!\!\!& &\!\!\!\!\!\!\!\! {} [N_{A_x} N_{A_y} N_{A_z}]\left[
                            \begin{array}{c}
                            a_x+u(b_x-a_x)+v(c_x-a_x) \\
                            a_y+u(b_y-a_y)+v(c_y-a_y) \\
                            a_z+u(b_z-a_z)+v(c_z-a_z) \\
                            \end{array}
                    \right] dv du \nonumber \\
  & & {}
\end{eqnarray}

Using the trinomial expansion
\begin{equation}
(x+y+z)^n = \sum_{k=0}^n\sum_{l=0}^{n-k}\left(
                                          \begin{array}{c}
                                            n \\
                                            k \\
                                          \end{array}
                                        \right)\left(
                                                 \begin{array}{c}
                                                   n-k \\
                                                   l \\
                                                 \end{array}
                                               \right)
                                               x^{n-l-k}y^{l}z^{k}
\end{equation}

The integral then reduces to
\begin{eqnarray}
  L_{s}(p) \!\!\!\!& = &\!\!\!\! \frac{- k_s \;\! I}{d_p^{(sh+3)/2}}\int_{u=u_{0}}^{u_1}\int_{v=v_{0}}^{v_1}
  \sum_{k=0}^{sh}\sum_{l=0}^{sh-k} \nonumber \\
  \!\!\!\!\!\!\!\!& &\!\!\!\!\!\!\!\! {}\left(
                                          \begin{array}{c}
                                            sh \\
                                            k \\
                                          \end{array}
                                        \right)\left(
                                                 \begin{array}{c}
                                                   sh-k \\
                                                   l \\
                                                 \end{array}
                                               \right)
                                               a_z^{sh-l-k}(b_z-a_z)^{l}(c_z-a_z)^{k}u^{l}v^{k} \nonumber\\
  \!\!\!\!\!\!\!\!& &\!\!\!\!\!\!\!\! {} [N_{A_x} N_{A_y} N_{A_z}]\left[
                            \begin{array}{c}
                            a_x+u(b_x-a_x)+v(c_x-a_x) \\
                            a_y+u(b_y-a_y)+v(c_y-a_y) \\
                            a_z+u(b_z-a_z)+v(c_z-a_z) \\
                            \end{array}
                    \right] dv du \nonumber \\
  & & {}
\end{eqnarray}

Rearranging the terms, the integral finally becomes
\begin{eqnarray}
 L_{s}(p) \!\!\!\!\!& = &\!\!\!\!\! \frac{- k_s \;\! I}{d_p^{(sh+3)/2}}\sum_{k=0}^{sh}\sum_{l=0}^{sh-k} \nonumber \\
 \!\!\!\!\!\!\!\!& &\!\!\!\!\!\!\!\! {} \left(\!\!\!\!
                                          \begin{array}{c}
                                            sh \\
                                            k \\
                                          \end{array}
                                        \!\!\!\!\right)\left(\!\!\!\!
                                                 \begin{array}{c}
                                                   sh-k \\
                                                   l \\
                                                 \end{array}
                                              \!\!\!\! \right)
                                               a_z^{sh-l-k}(b_z-a_z)^{l}(c_z-a_z)^{k}u^{l}v^{k}\nonumber\\
  \!\!\!\!\!\!\!\!& &\!\!\!\!\!\!\!\! {} \int_{u=u_{0}}^{u_1}\int_{v=v_{0}}^{v_1}a_0u^{l}v^{k} +
  a_1u^{l+1}v^{k} + a_2u^{l}v^{k+1} dv du\nonumber
\end{eqnarray}
\begin{eqnarray}
  a_{0} & = & \vec{N_A}.\vec{a} \nonumber \\
  a_{1} & = & \vec{N_A}.(\vec{b}-\vec{a}) \nonumber \\
  a_{2} & = & \vec{N_A}.(\vec{c}-\vec{a})
\end{eqnarray}
which is a sum of integrals of polynomials that can be again readily
evaluated in closed-form achieving a complexity of \emph{O}($sh^2$)
for Phong-like materials.

\subsection{Non-Constant Area Light Sources}
In the previous sections, we derived closed-form solutions for
rendering various types of materials lit by a constant area light
source of rectangular shape. These solutions reduce to a point spot
light or a linear spot light source by simple dimensionality
reduction. Therefore, our close-form solutions nicely include other
popular light source models. To demonstrate that our approach is
general, we now show that it can also be equally extended to provide
a solution for direct lighting in scenes lit by environment
cubemaps. Indeed, from the point of view presented in this paper,
each side of a cubemap is simply an area light source with pointwise
varying color and intensity. Therefore, we would simply require to
extend the results in the previous sections to the case of
non-constant area light sources.

We demonstrate the basic idea for Lambertian materials, which can
also be extended in a similar manner as before to other types of
material. For Lambertian materials the light integral becomes
\begin{equation}
 L_{d}(p) = k_d
 \int_{Area}I_{A_{}}max(0,\cos\theta_i)\frac{\cos\theta_A}{r_A^2}dA
\end{equation}
Following the same steps as before the integral can be simplified to
\begin{eqnarray*}
  L_{d}(p) & = & \frac{-k_d}{d^2}\int_{u=u_{0}}^{u_1}\int_{v=v_{0}}^{v_1}I(u,v)(l_{00}+l_{01}v+l_{02}v^2+\nonumber\\
  & & {} l_{10}u+l_{11}uv+l_{20}u^2) dv du \label{eq:rend}
\end{eqnarray*}
%\vspace*{-0.8cm}
\begin{eqnarray}
  l_{00}\!\!\!\!\! & = & \!\!\!\!\!a_z \times (\vec{a}.\vec{N_A}) \nonumber \\
  l_{01}\!\!\!\!\! & = & \!\!\!\!\!a_z \times ((\vec{c}-\vec{a}).\vec{N_A})+(c_z-a_z)\times (\vec{a}.\vec{N_A}) \nonumber \\
  l_{02}\!\!\!\!\! & = & \!\!\!\!\!(c_z-a_z) \times ((\vec{c}-\vec{a}).\vec{N_A}) \nonumber \\
  l_{10}\!\!\!\!\! & = & \!\!\!\!\!a_z \times ((\vec{b}-\vec{a}).\vec{N_A})+(b_z-a_z)\times (\vec{a}.\vec{N_A}) \nonumber \\
  l_{11}\!\!\!\!\! & = & \!\!\!\!\!(c_z-a_z) \times ((\vec{c}-\vec{a}).\vec{N_A}) + (c_z-a_z) \times ((\vec{b}-\vec{a}).\vec{N_A}) \nonumber \\
  l_{20}\!\!\!\!\! & = & \!\!\!\!\!(b_z-a_z) \times ((\vec{b}-\vec{a}).\vec{N_A})
\end{eqnarray}
where $I(u,v)$ is the varying light color and intensity, which is
essentially a pixel in an image of a cubemap.

This integral can be solved in closed-form only if we can provide a
closed-form expression for the pixel value $I(u,v)$. A natural way
to do this would be to use some basis function. In principle, any
basis may be used to compute $I(u,v)$. However, we used the Discrete
Cosine Transform (DCT), since it provides two advantages. First,
combined with the light integral equation, it lends itself to a
closed-form solution, second DCT is a well established and widely
used compression tool.  Therefore $I(u,v)$ is given by an inverse
cosine transform of a set of coefficients precomputed by
preprocessing the image in the cubemap, as follows:
\begin{eqnarray*}
I(u,v)=\sum_{i=0}^{N-1}\sum_{j=0}^{M-1}\alpha_i\alpha_jC_{ij}\cos(k_{i0}u+k_{i1})\cos(k_{j0}v+k_{j1})
\label{eq:DCTpixel}
\end{eqnarray*}
\begin{eqnarray}
\mbox{where}\;\;\;\; \alpha_i &=& \left\{
                    \begin{array}{l}
                      1/\sqrt{N},i=0 \\
                      \sqrt{2/N},1 \leq i \leq N-1 \\
                    \end{array}
                  \right.\nonumber\\
%\end{eqnarray*}
%\begin{eqnarray*}
\alpha_j &=& \left\{
                    \begin{array}{l}
                      1/\sqrt{M},j=0 \\
                      \sqrt{2/M},1 \leq j \leq M-1 \\
                    \end{array}
                  \right.\nonumber\\
%\end{eqnarray*}
%\begin{eqnarray}
  k_{i0}&=&\pi i (N-1)/N, \;\;\;\; k_{j0}=\pi j (M-1)/M \nonumber\\
  k_{i1}&=&\pi i/2N,      \;\;\;\;\;\;\;\;\;\;\;\;\;\;\; k_{j1}=\pi j/2M
\end{eqnarray}

The $C_{ij}$'s are the DCT coefficients.  The coefficients are
precomputed for each color channel and are stored as a cubemap that
is used as an input to our rendering equation in (\ref{eq:rend}).
The preprocessing step requires $N_c\log (N_c)$ time, if we use
$N_c$ coefficients.

%An example of a DCT cubemap of one of the color channels is shown in
%Figure \ref{fig:DCTmap}.
%\begin{figure}[h]
%\centering
%\includegraphics[width=2in]{dccmap.pdf}
%\caption{An example of a DCT cubemap precomputed for rendering in
%our
%  method.} \label{fig:DCTmap}
%\end{figure}

Upon substituting from (\ref{eq:DCTpixel}) into (\ref{eq:rend}), our
integral becomes
\begin{eqnarray}
 L_{d}(p) \!\!\!\!\!\!\!\!\!\!&& =
  \frac{-k_d}{d^2}\int_{u=u_{0}}^{u_1}\int_{v=v_{0}}^{v_1}\sum_{i=0}^{N-1}\sum_{j=0}^{M-1}\alpha_i\alpha_jC_{ij}\nonumber\\
  \!\!\!\!\!\!\!\!\!\!\!\!& & {} \cos(k_{i0}u+k_{i1})\cos(k_{j0}v+k_{j1})\nonumber\\
  \!\!\!\!\!\!\!\!\!\!\!\!& & {} (l_{00}+l_{01}v+l_{02}v^2+l_{10}u+l_{11}uv+l_{20}u^2) dv
  du\;\;\;\;\;\;\;\;\;
\end{eqnarray}

We then rearrange the integral to get the following
\begin{eqnarray}
  L_{d}(p) \!\!\!\!& = &\!\!\!\!
  \frac{-k_d}{d^2}\sum_{i=0}^{N-1}\sum_{j=0}^{M-1}\alpha_i\alpha_jC_{ij}\int_{u=u_{0}}^{u_1}\int_{v=v_{0}}^{v_1}\nonumber\\
  \!\!\!\!\!\!\!\!& &\!\!\!\!\!\!\!\! {} \cos(k_{i0}u+k_{i1})\cos(k_{j0}v+k_{j1})\nonumber\\
  \!\!\!\!\!\!\!\!& &\!\!\!\!\!\!\!\! {} (l_{00}+l_{01}v+l_{02}v^2+l_{10}u+l_{11}uv+l_{20}u^2) dv du \;\;\;\;\;\;\;\;\;
\end{eqnarray}

A closed-form solution can now be found for the integral part of the
above equation, which we denote by $\Im_{ij}$. The closed form
generated will have divisions by the constants $k_{i0}$, $k_{i1}$,
$k_{j0}$ and $k_{j1}$. However, at i=0 and j=0 these constants
evaluate to zero, leading to divisions by zero. To solve this
problem, the integral is divided into four cases:

\begin{enumerate}
  \item Case 1: $i\neq 0$ and $j\neq 0$
  \begin{eqnarray}
  I_{ij} \!\!\!\!\!\!& = &\!\!\!\!\!\!
  \int_{u=u_{0}}^{u_1}\int_{v=v_{0}}^{v_1}\cos(k_{i0}u+k_{i1})\cos(k_{j0}v+k_{j1})\nonumber\\
  \!\!\!\!\!\!\!\!& &\!\!\!\!\!\!\!\! (l_{00}+l_{01}v+l_{02}v^2+l_{10}u+l_{11}uv+l_{20}u^2) dv
  du \;\;\;\;\;\;\;\;\;
\end{eqnarray}
  \item Case 2: $i = 0$ and $j\neq 0$
  \begin{eqnarray}
  I_{0j} \!\!\!\!\!\!& = &\!\!\!\!\!\!
  \int_{u=u_{0}}^{u_1}\int_{v=v_{0}}^{v_1}\cos(k_{j0}v+k_{j1})\nonumber\\
  \!\!\!\!\!\!\!\!& &\!\!\!\!\!\!\!\! (l_{00}+l_{01}v+l_{02}v^2+l_{10}u+l_{11}uv+l_{20}u^2) dv
  du \;\;\;\;\;\;\;\;\;
\end{eqnarray}
  \item Case 3: $i\neq 0$ and $j = 0$
  \begin{eqnarray}
  I_{i0} \!\!\!\!\!\!& = &\!\!\!\!\!\!
  \int_{u=u_{0}}^{u_1}\int_{v=v_{0}}^{v_1}\cos(k_{i0}u+k_{i1})\nonumber\\
  \!\!\!\!\!\!\!\!& &\!\!\!\!\!\!\!\! (l_{00}+l_{01}v+l_{02}v^2+l_{10}u+l_{11}uv+l_{20}u^2) dv
  du \;\;\;\;\;\;\;\;\;
\end{eqnarray}
  \item Case 4: $i=0$ and $j=0$
  \begin{eqnarray}
  I_{00} \!\!\!\!\!\!& = &\!\!\!\!\!\!
  \int_{u=u_{0}}^{u_1}\int_{v=v_{0}}^{v_1}\nonumber\\
  \!\!\!\!\!\!\!\!& &\!\!\!\!\!\!\!\!(l_{00}+l_{01}v+l_{02}v^2+l_{10}u+l_{11}uv+l_{20}u^2) dv
  du \;\;\;\;\;\;\;\;\;
\end{eqnarray}
which is the same as the constant area light source for lambertian
materials case.
\end{enumerate}

The above integrals can now be evaluated using the bounds in table-2
and the color value at a point $p$ on a Lambertian surface will be
\begin{eqnarray}
  L_{d}(p) & = &
  \frac{-k_d}{d^2}\sum_{i=0}^{N-1}\sum_{j=0}^{M-1}\alpha_i\alpha_jC_{ij}\Im_{ij}
\end{eqnarray}
%\vspace*{-0.3cm} 
where, 
%\vspace*{-0.3cm}
\begin{eqnarray*}
\Im_{ij} & = & \left\{
                    \begin{array}{c}
                        I_{ij}, \hspace{.1in} i \neq 0_{} \hspace{.1in} \& \hspace{.1in} j \neq 0 \\
                        I_{0j}, \hspace{.1in} i=0_{} \hspace{.1in} \& \hspace{.1in} j \neq 0 \\
                        I_{i0}, \hspace{.1in} i \neq 0_{} \hspace{.1in} \& \hspace{.1in} j=0 \\
                        I_{00}, \hspace{.1in} i=0_{} \hspace{.1in} \& \hspace{.1in} j=0
                    \end{array}
                    \right.
\end{eqnarray*}

The number of coefficients generated by the discrete cosine
transform of the image is equal to the image resolution $M \times
N$.  If desired, all $M \times N$ coefficients can be used to
compute $L_o(p)$ in \emph{O}(MN) time, or alternatively we can
truncate the coefficients to some desired cut-off low frequency
values trading between accuracy and speed.

The nature of our closed-form solution allows our algorithm to work
for high frequency as well as low frequency. Another significant
advantage is that we are able to achieve realistic shading using
only \emph{one} coefficient per face, reducing the number of
coefficients needed to achieve realistic shading as compared to
spherical harmonics.

If \emph{O}(1) complexity is desired, the analytic solution for the
lambertian surfaces under constant lighting can be used, such that
$I=C_{00}/\sqrt{MN}$.

\section{Results and Discussion}

\begin{figure}[t]

\centering
\includegraphics[width=9.5cm]{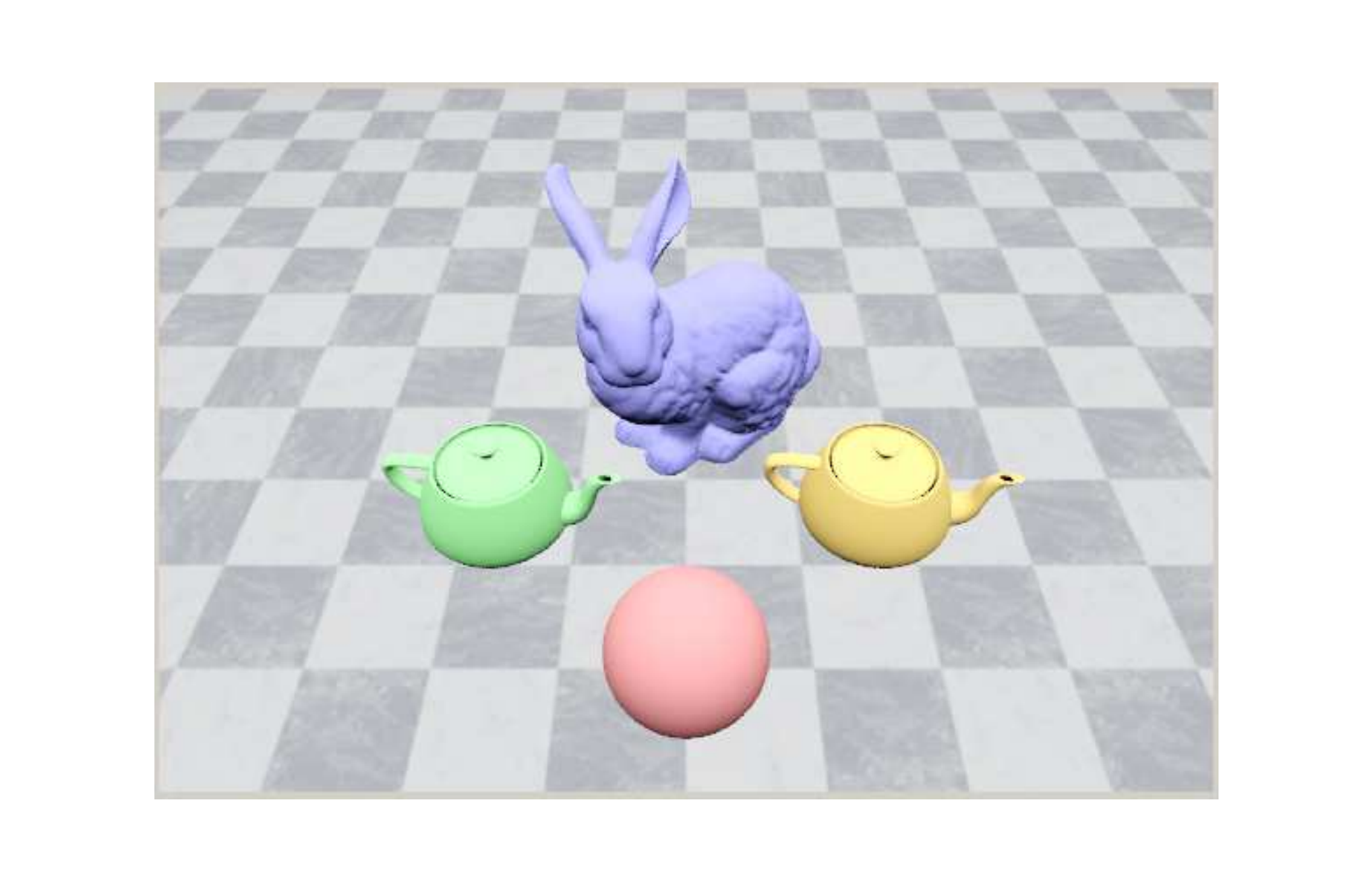}
\hspace*{-1.1cm}
\includegraphics[width=9.5cm]{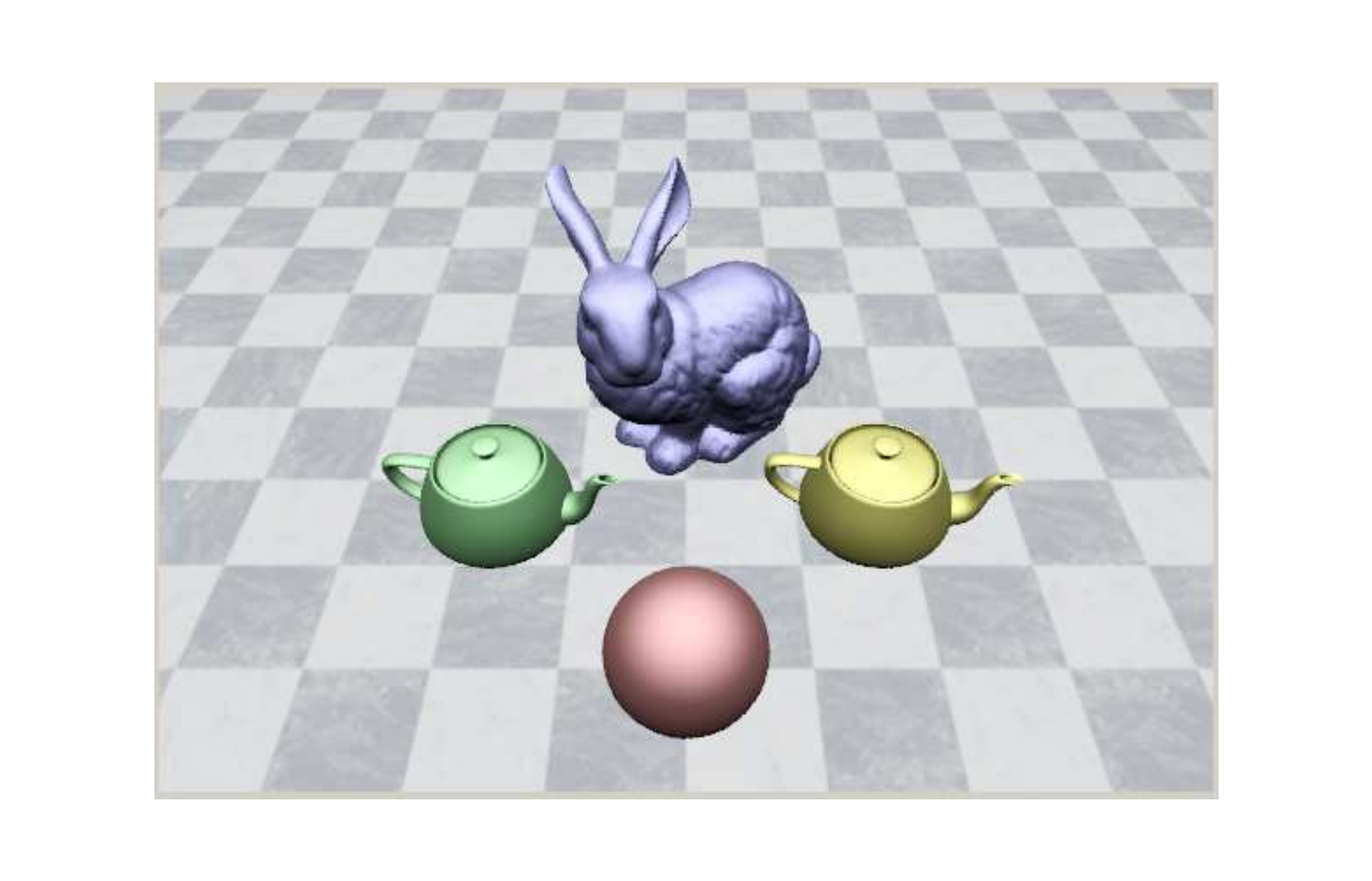}
%\vspace*{-5mm} 
\caption{Lambertian and Phong-like materials lit by a
constant area light source}\label{fig:scenes}
\end{figure}

\begin{figure*}[h]
\begin{center}
\begin{tabular}{cccc}
\hspace*{-2mm}\includegraphics[width=28mm]{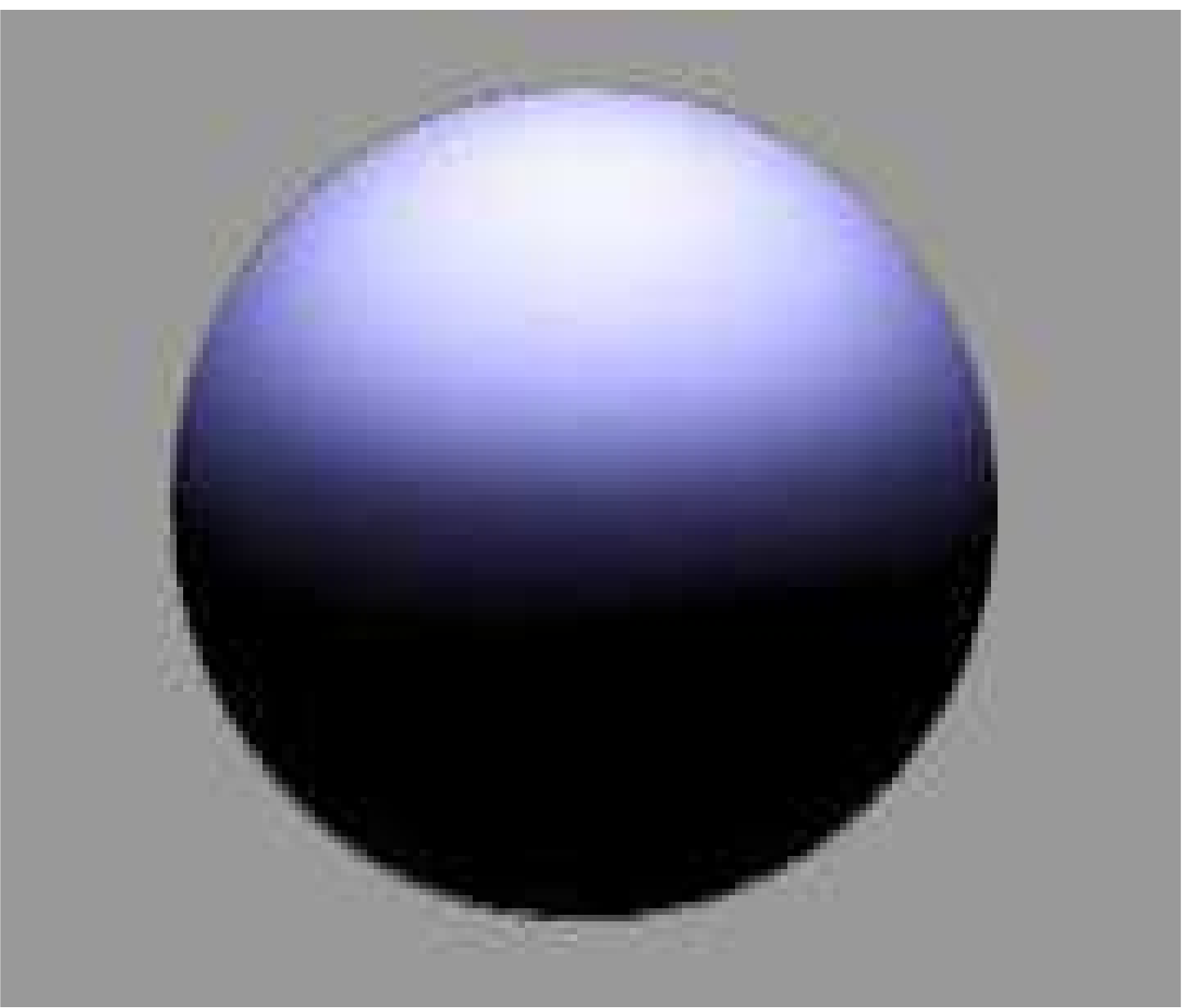}
&\hspace*{-3mm}
\includegraphics[width=28mm]{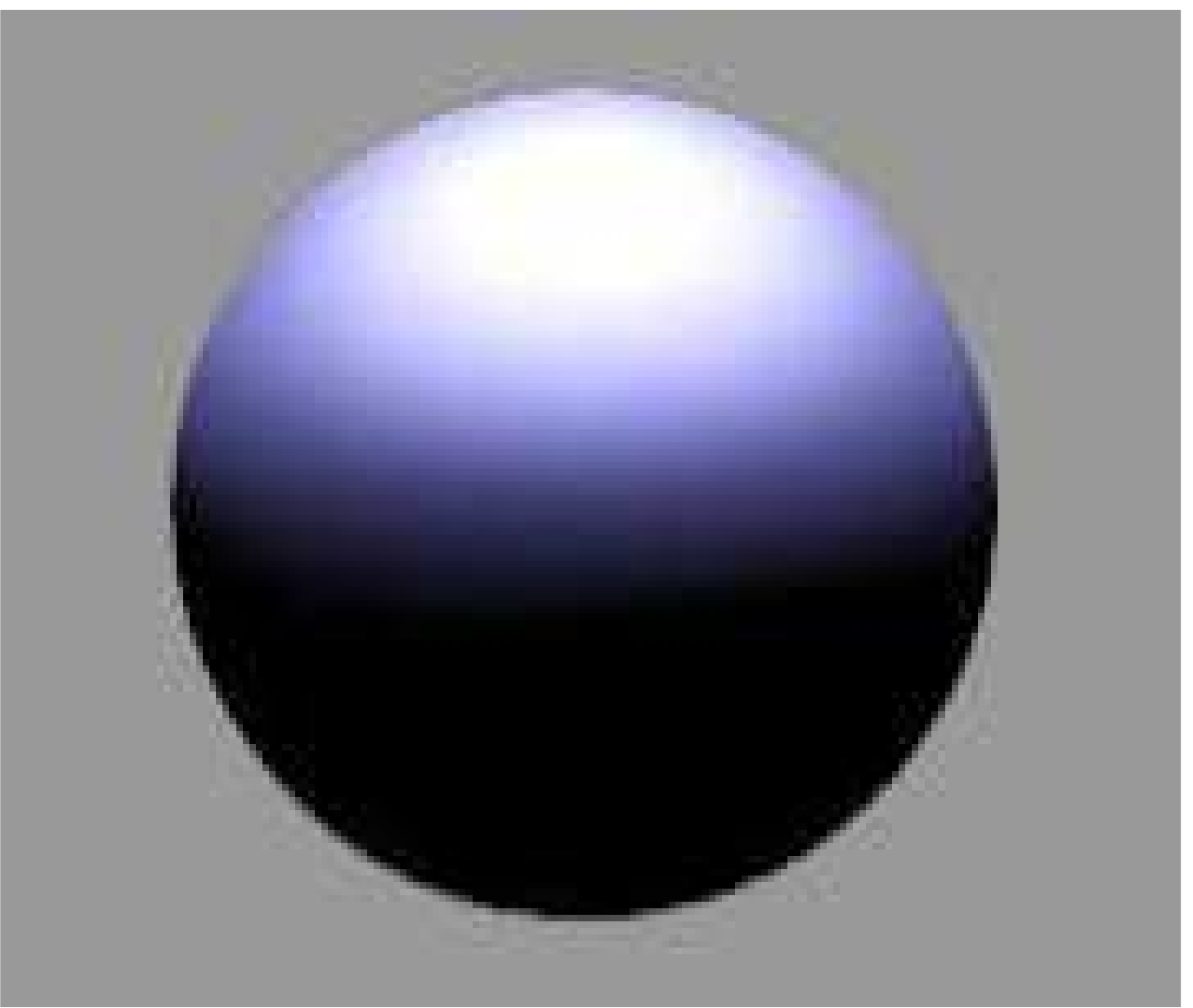} & \hspace*{-3mm}
\includegraphics[width=28mm]{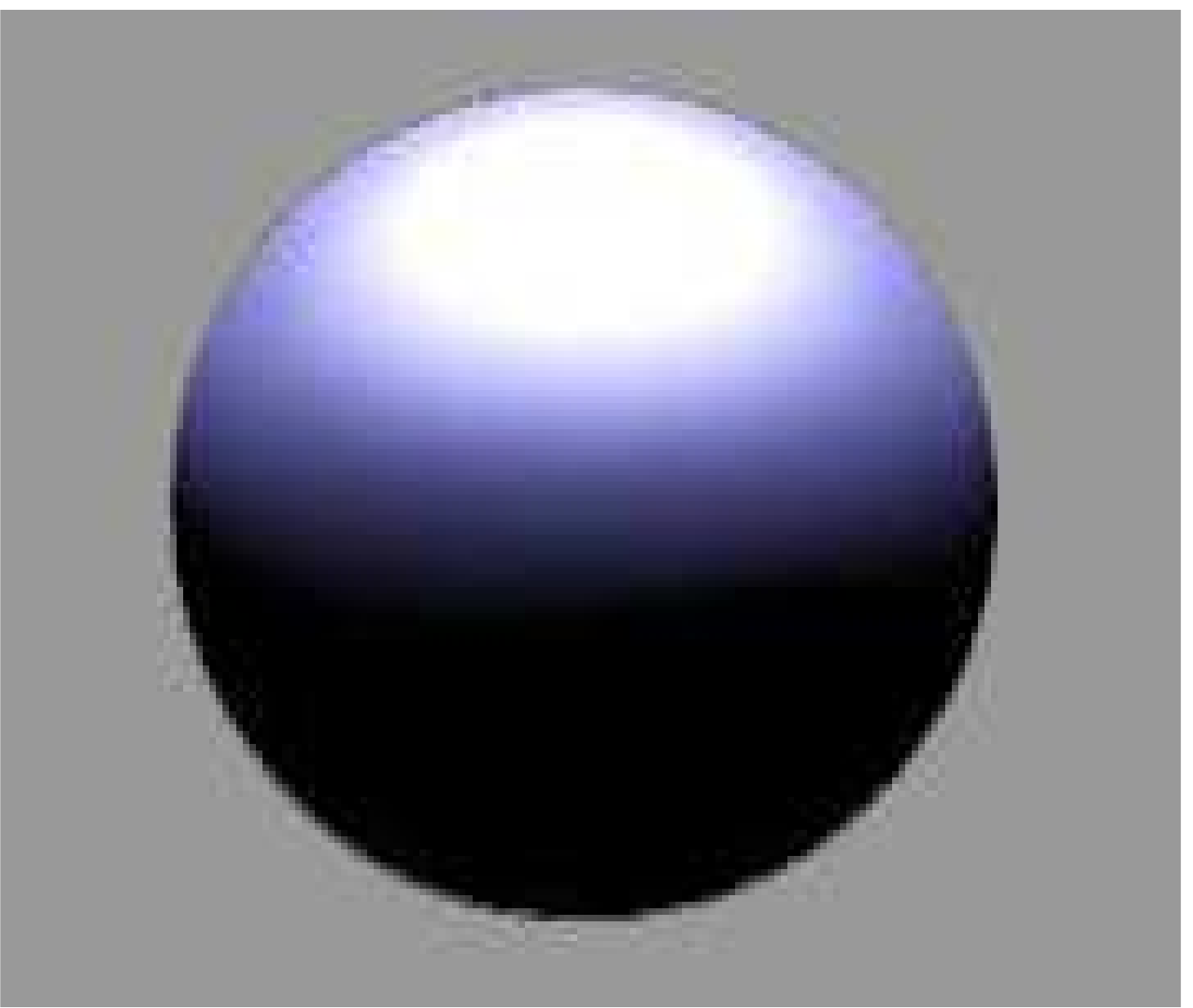} &\hspace*{-3mm}
\includegraphics[width=28mm]{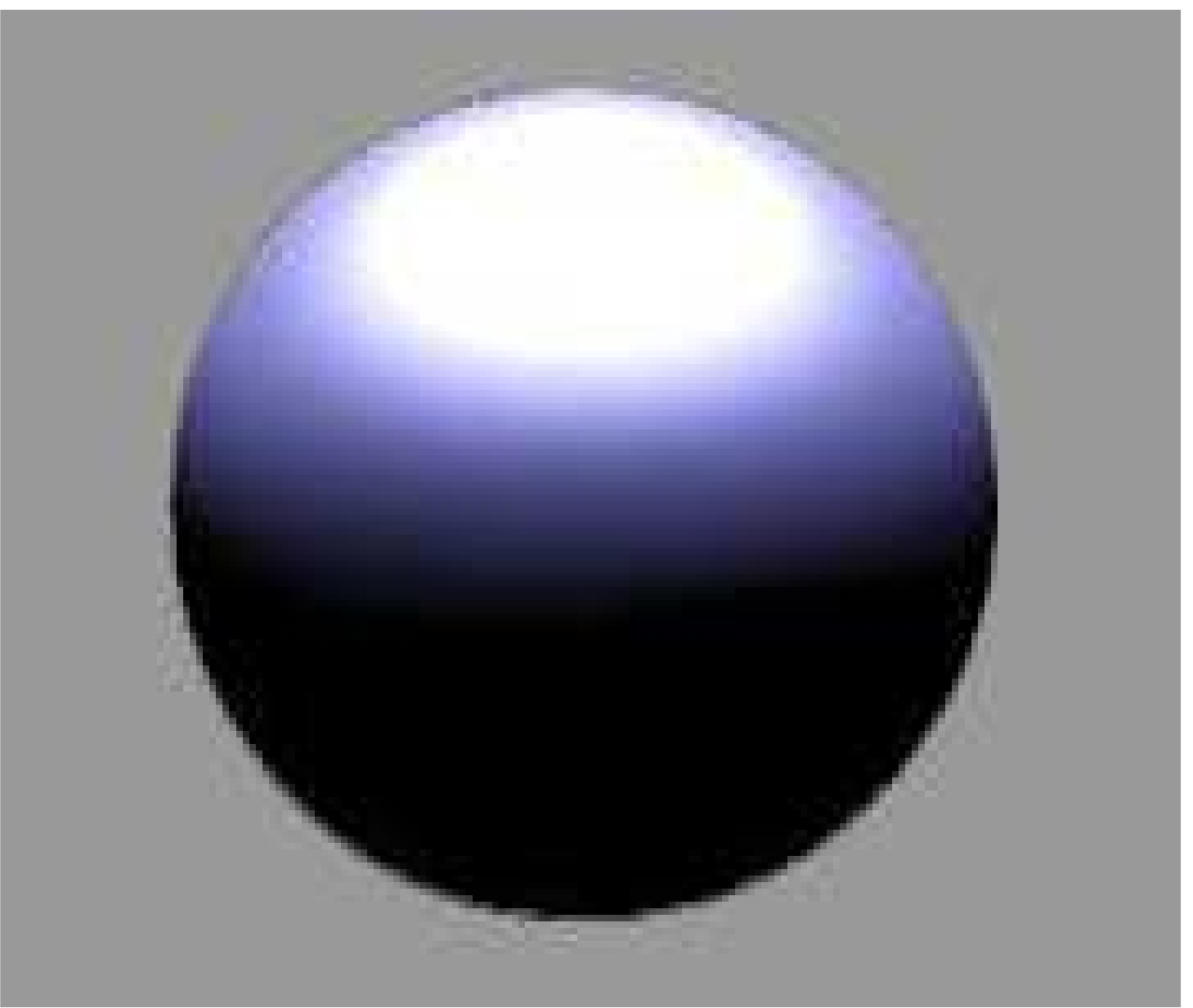} \\
$sh=2$ & $sh=3$ & $sh=4$ & $sh=5$ \\
\hspace*{-2mm}\includegraphics[width=28mm]{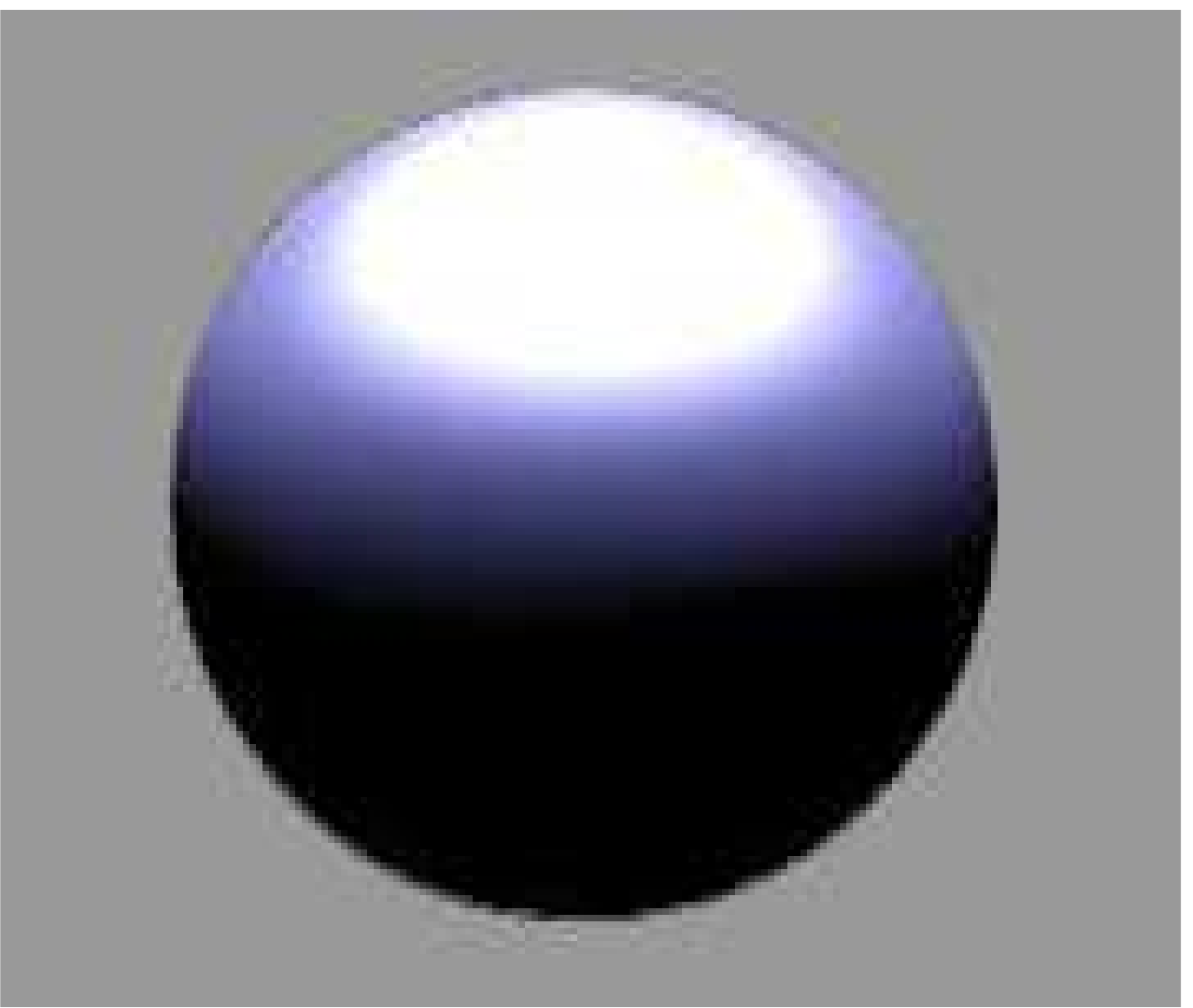}
&\hspace*{-3mm}
\includegraphics[width=28mm]{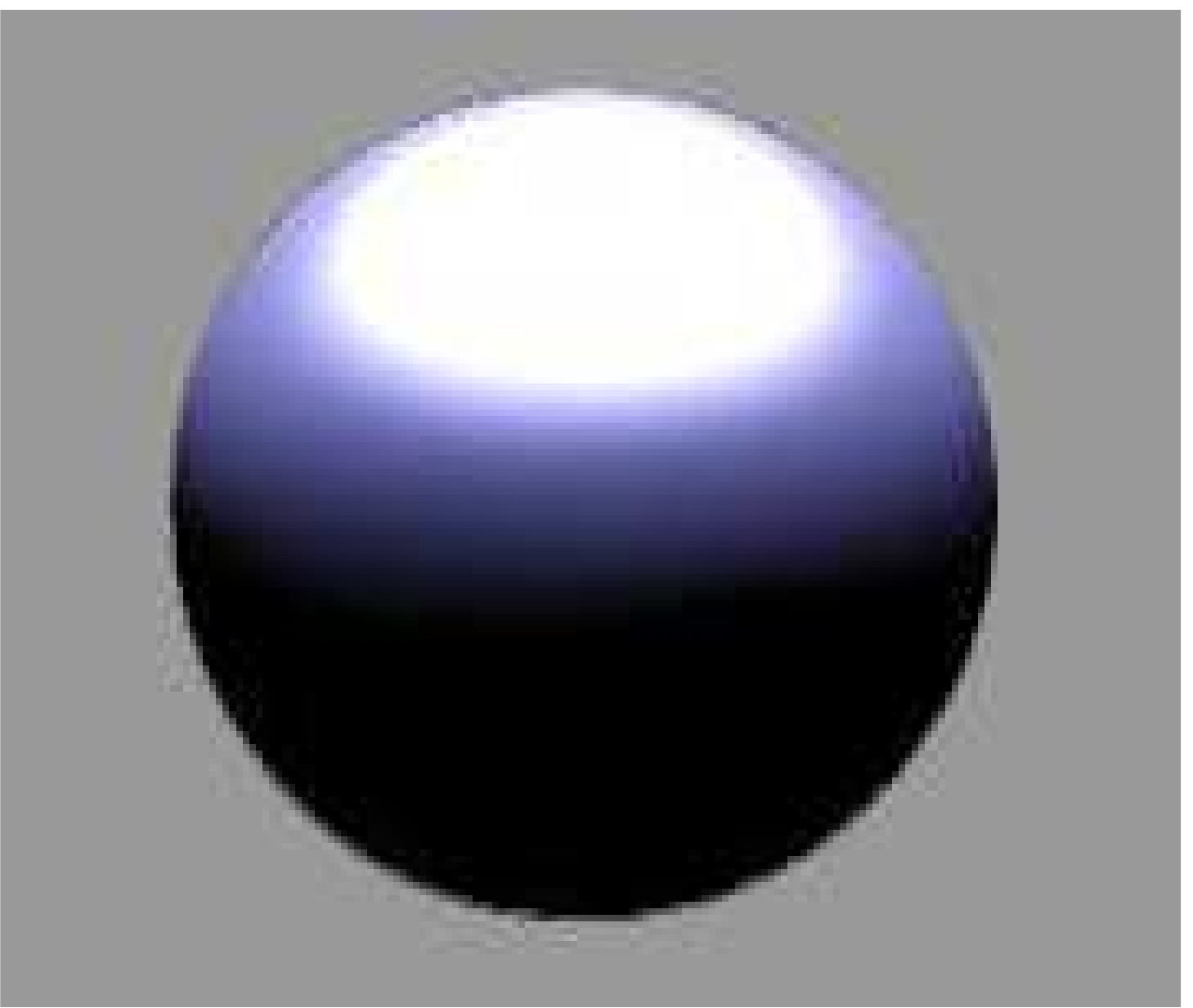} &\hspace*{-3mm}
\includegraphics[width=28mm]{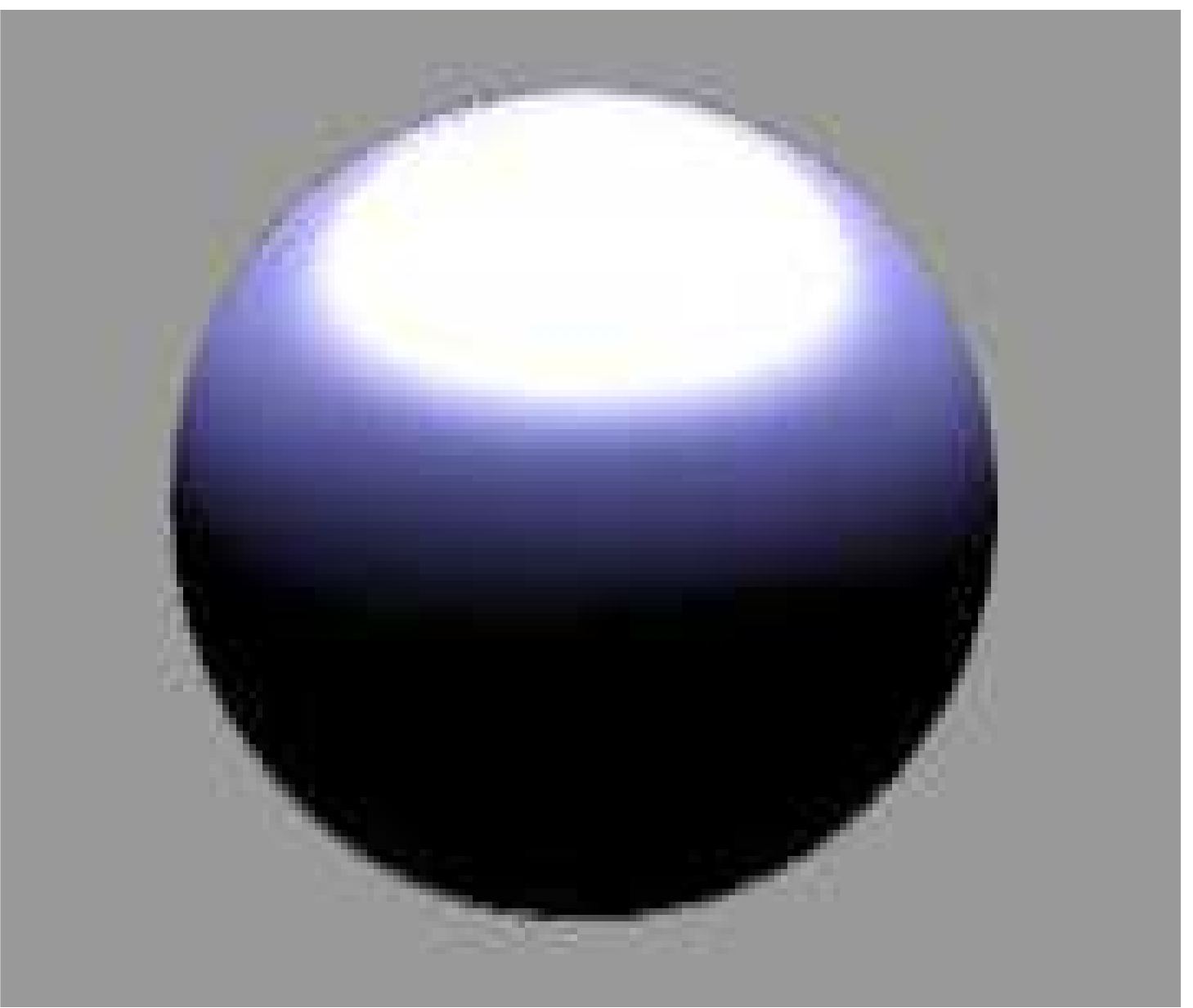} &\hspace*{-3mm}
\includegraphics[width=28mm]{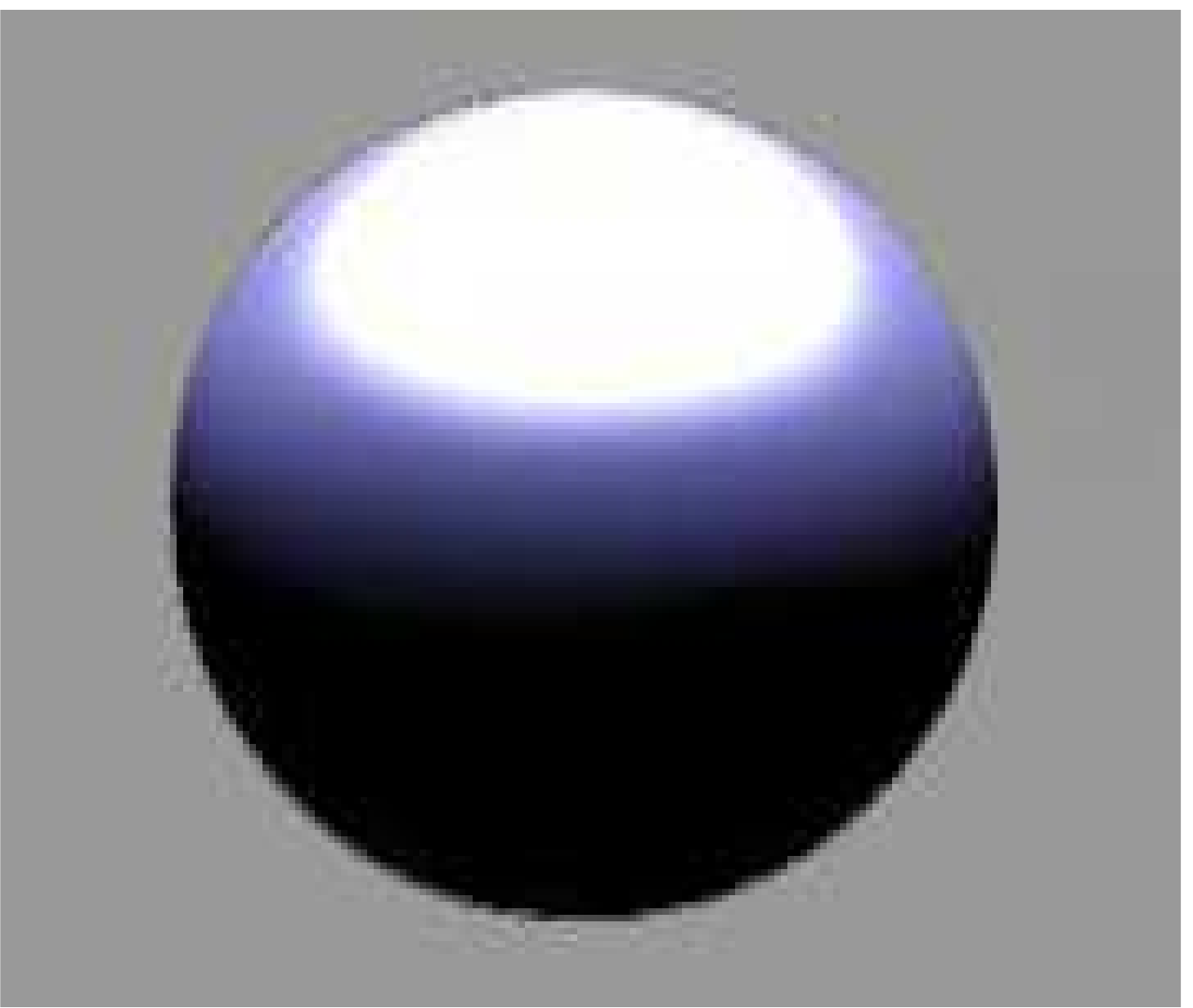} \\
$sh=6$ & $sh=7$ & $sh=8$ & $sh=9$ \\
\end{tabular}
\end{center}
\caption{Examples of rendering Phong-like material with different
  shininess parameters under an area light source.} \label{fig:phong}
%\end{figure}
%
%\begin{figure}[h]
\centering
\includegraphics[width=5cm]{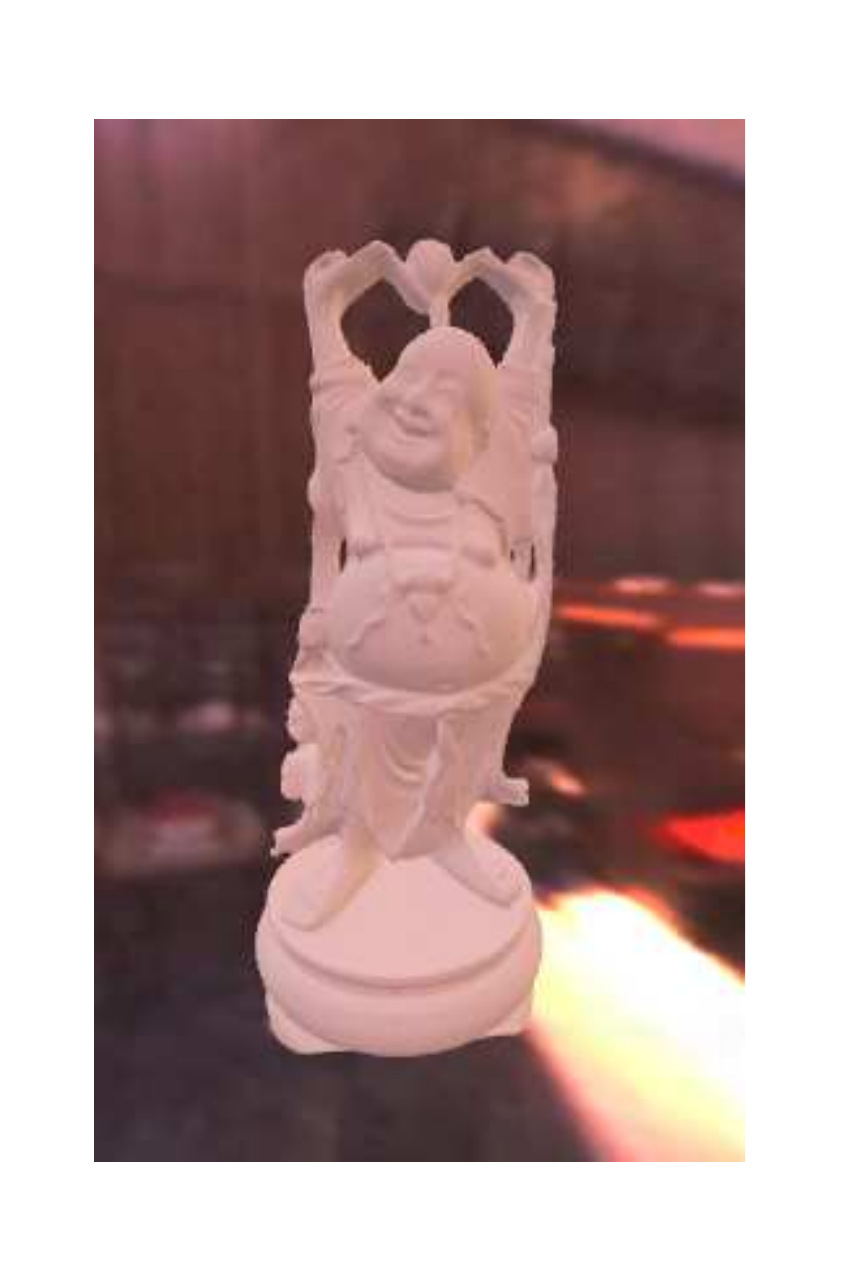}
\includegraphics[width=5cm]{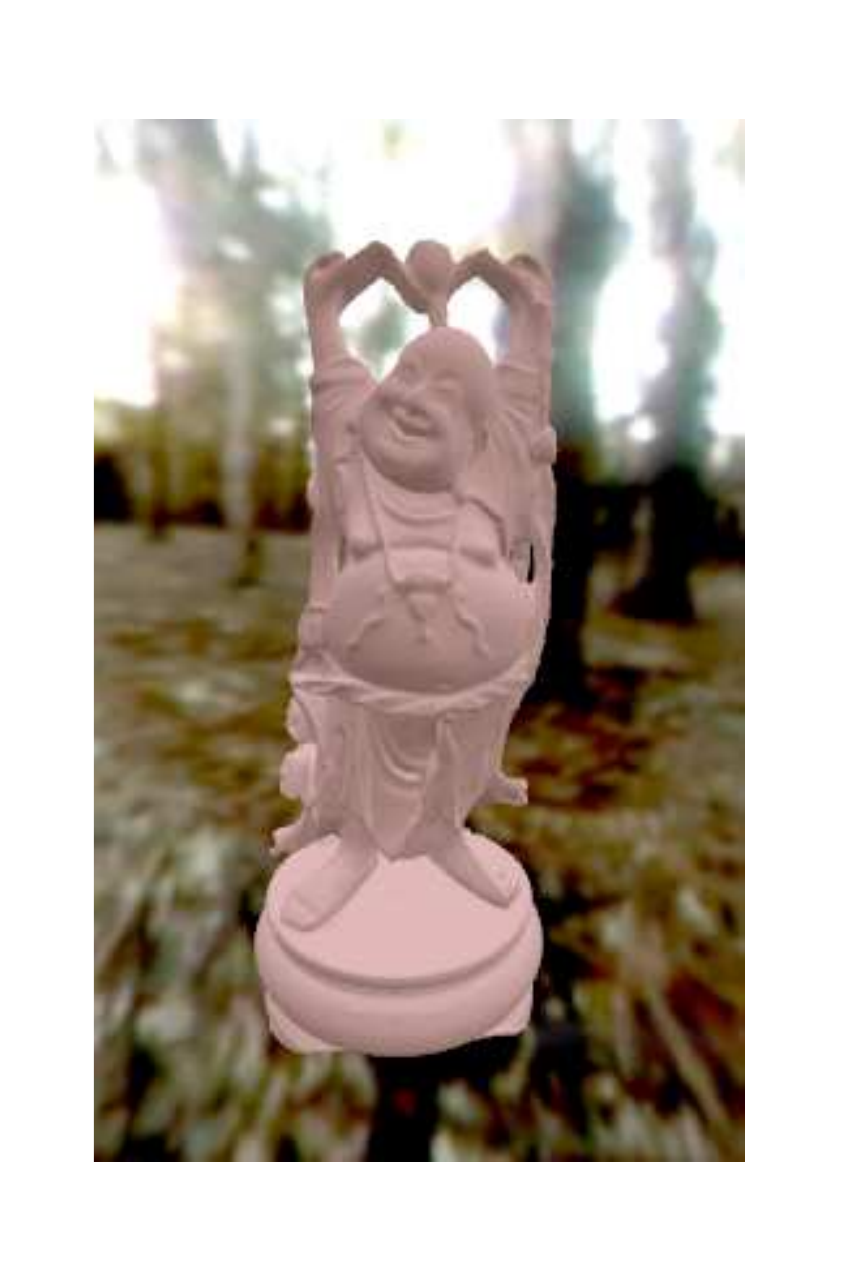}
\includegraphics[width=5cm]{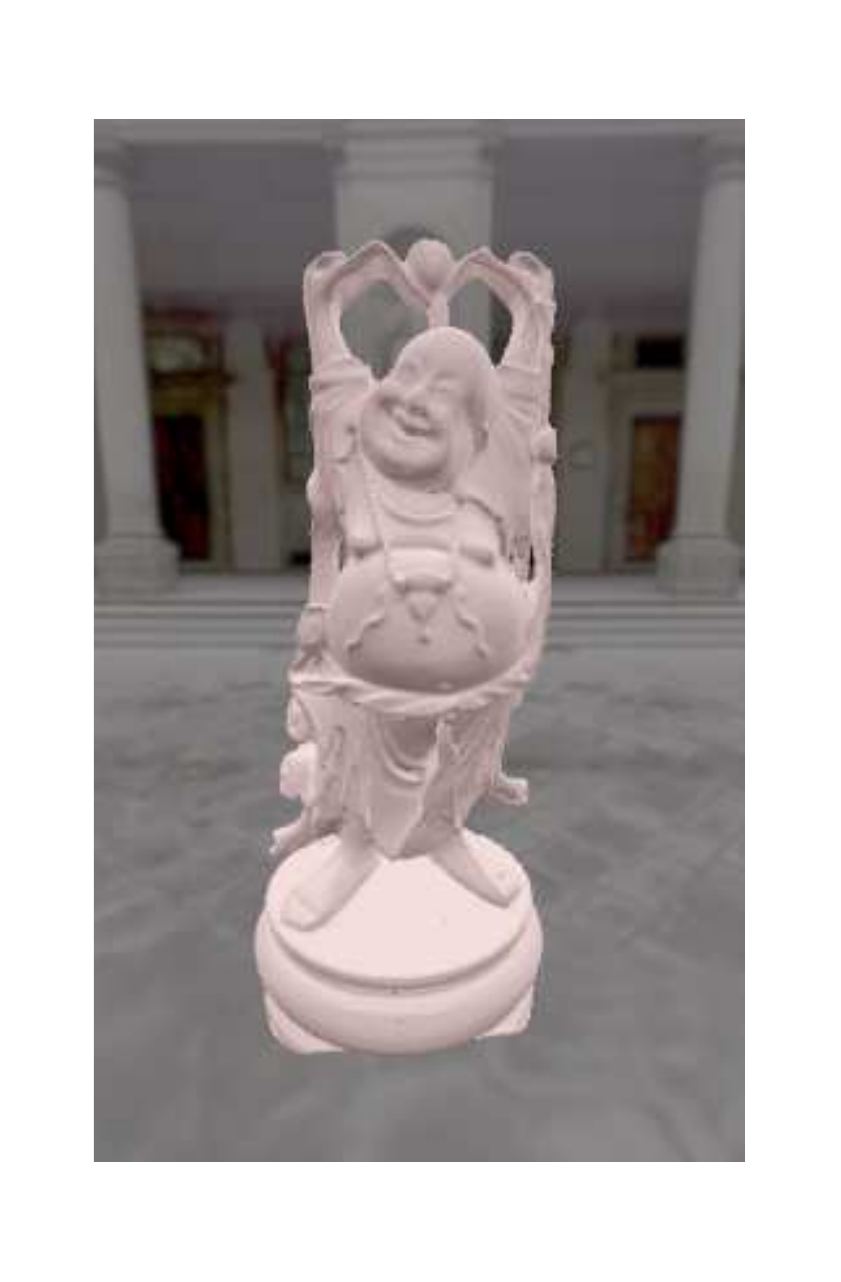}
%\vspace*{-0.7cm} 
\caption{Lambertian buddha lit by different
environments}\label{fig:buddha-env} 
%\vspace*{0.4cm}
%\end{figure}
%
%\begin{figure}[h]
\centering
\includegraphics[width=6.3cm]{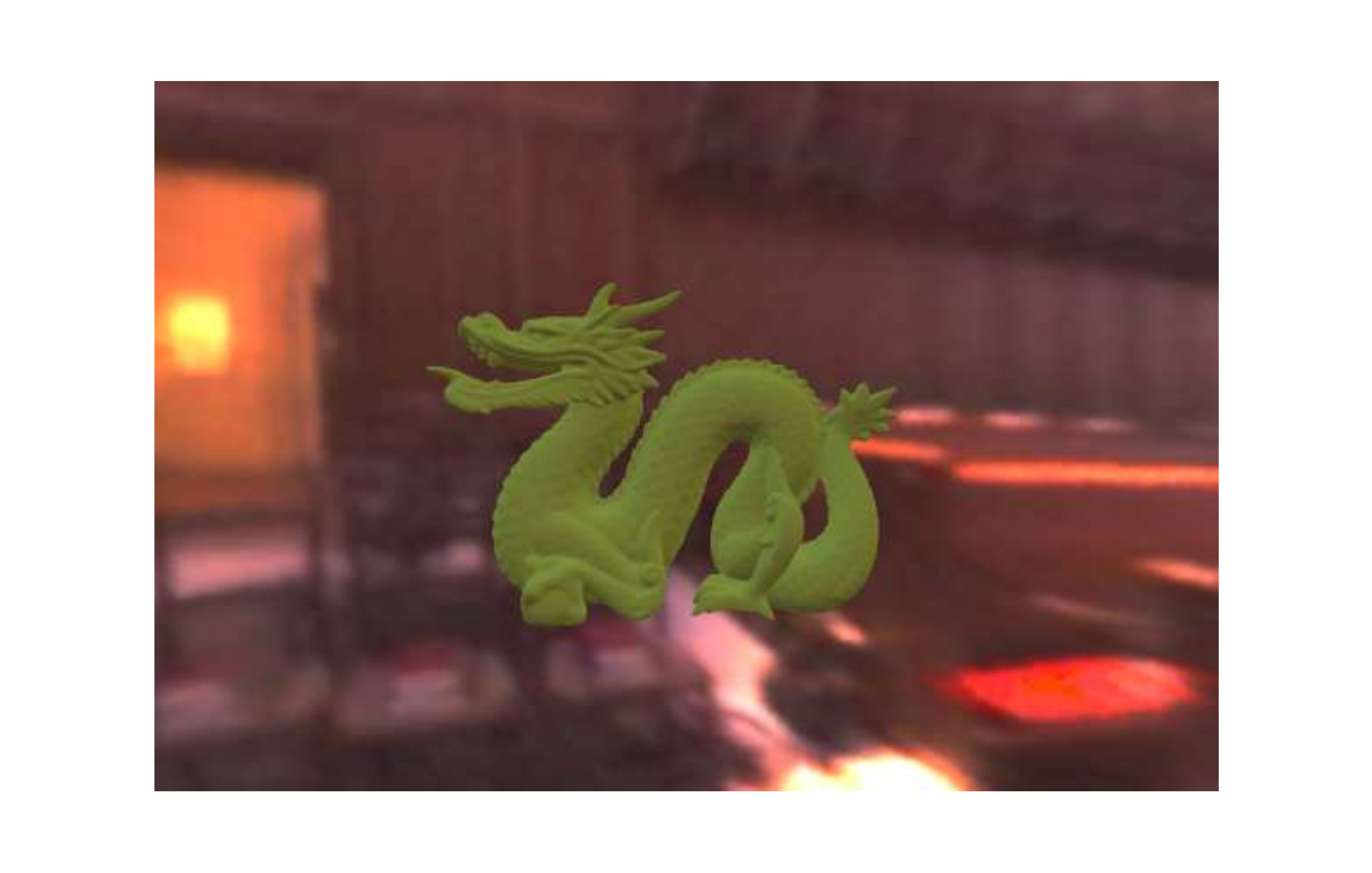}
\hspace*{-1.4cm} \includegraphics[width=6.3cm]{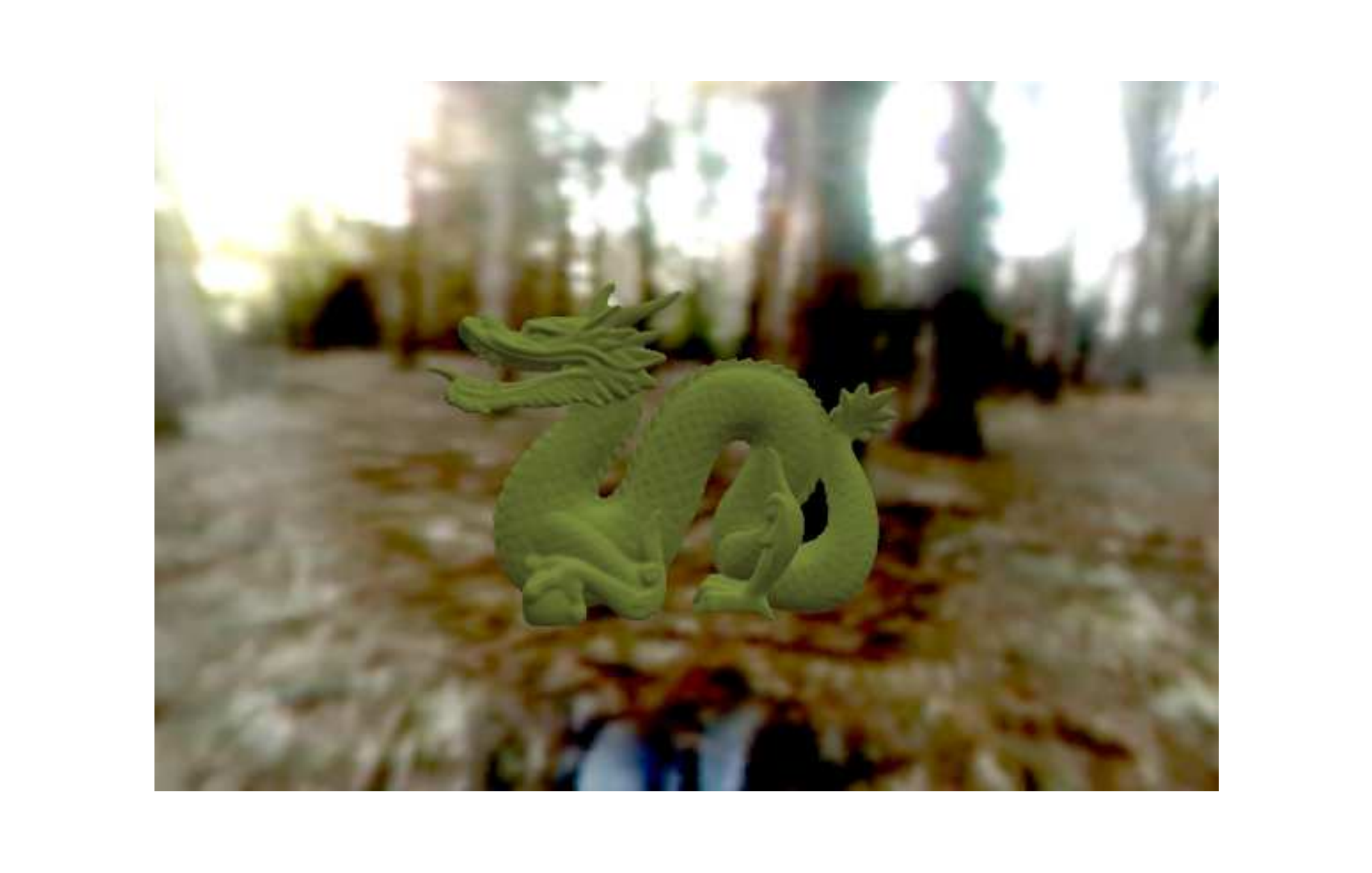}
\hspace*{-1.4cm} \includegraphics[width=6.3cm]{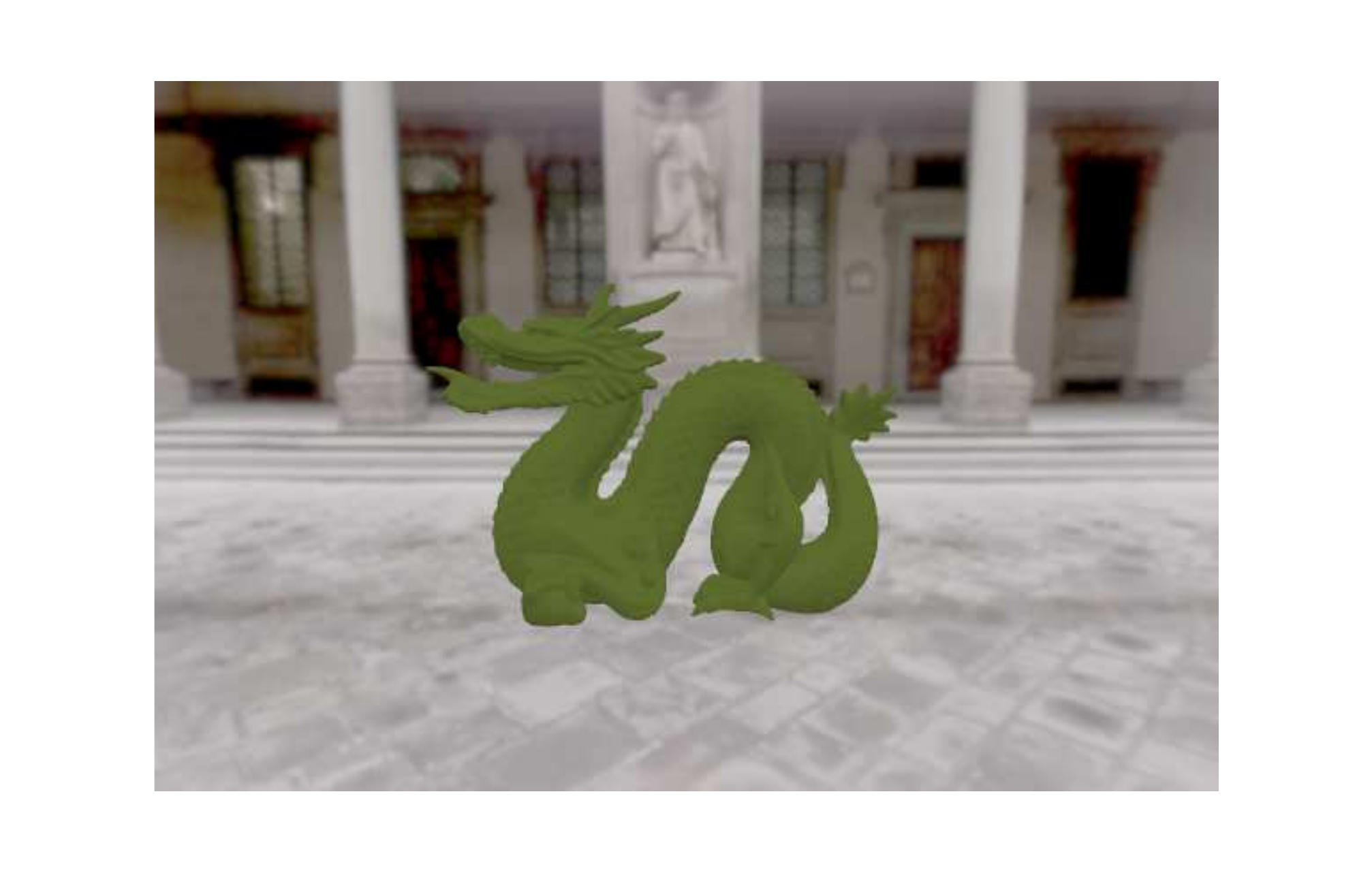}
%\vspace*{-0.4cm} 
\caption{Lambertian dragon lit by different
environments}\label{fig:dragon-env} 
%\vspace*{0.4cm}
%\end{figure}
%
%\begin{figure}[h]
\centering
\includegraphics[width=6.3cm]{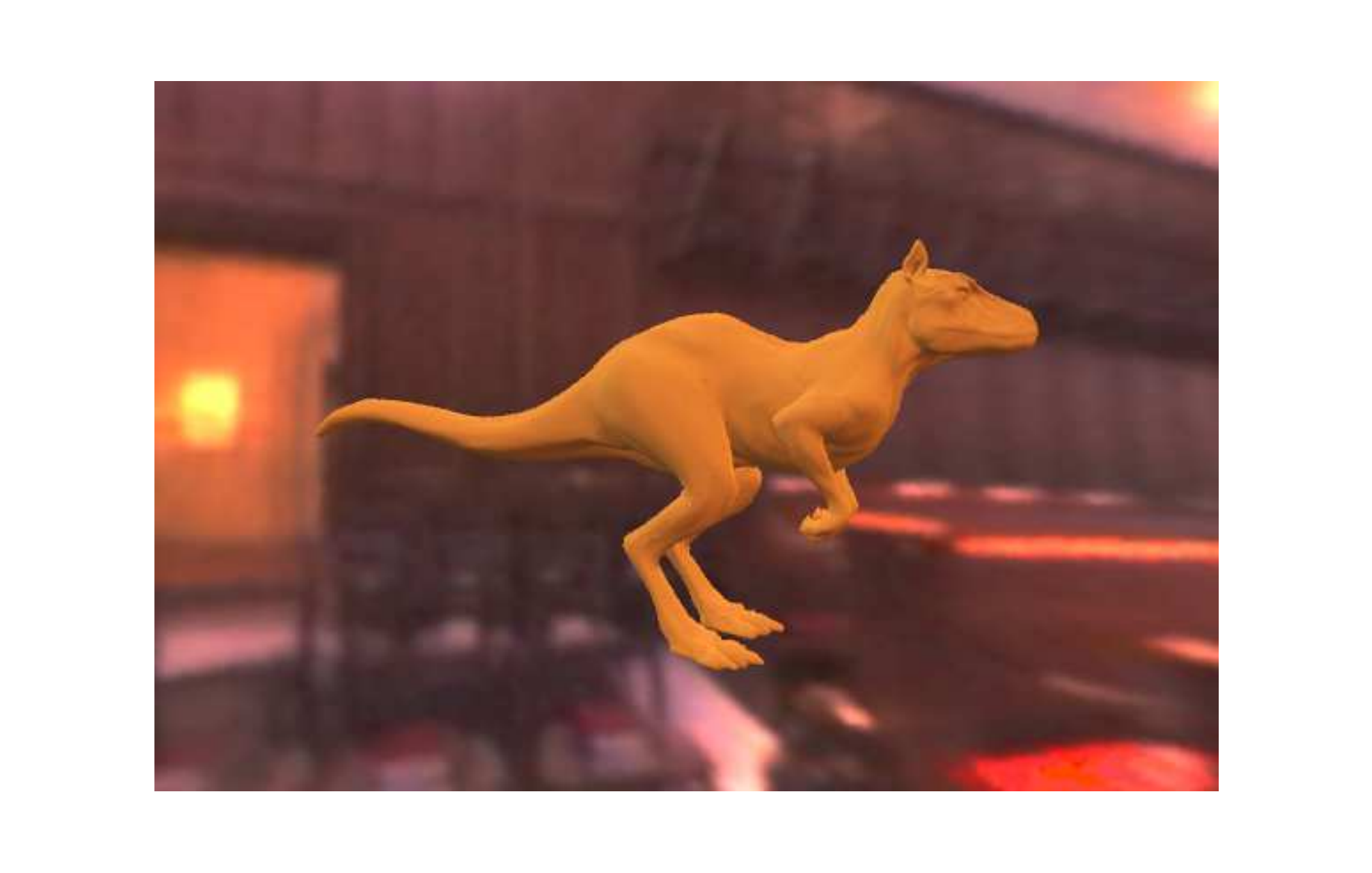}
\hspace*{-1.4cm} \includegraphics[width=6.3cm]{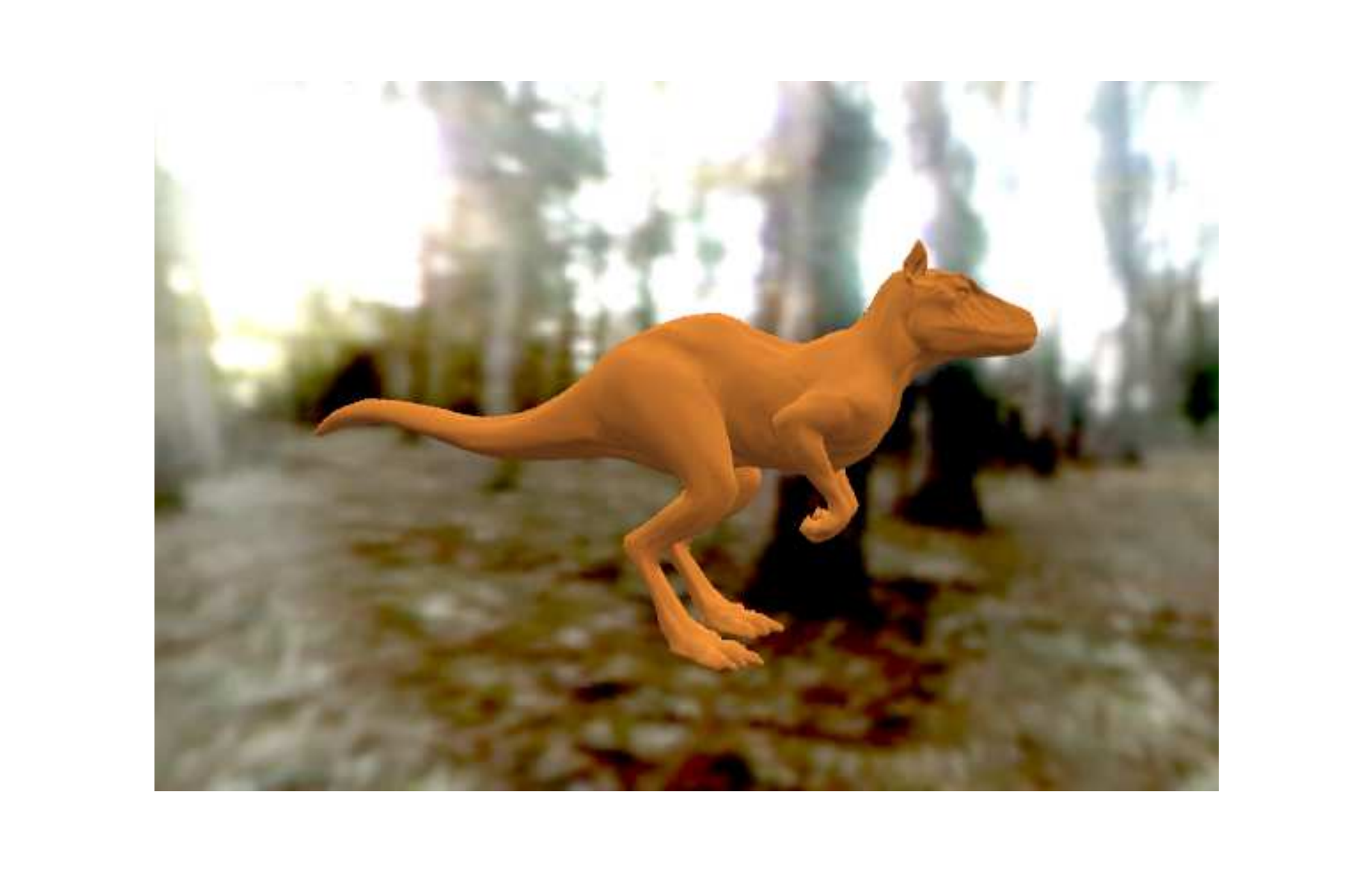}
\hspace*{-1.4cm} \includegraphics[width=6.3cm]{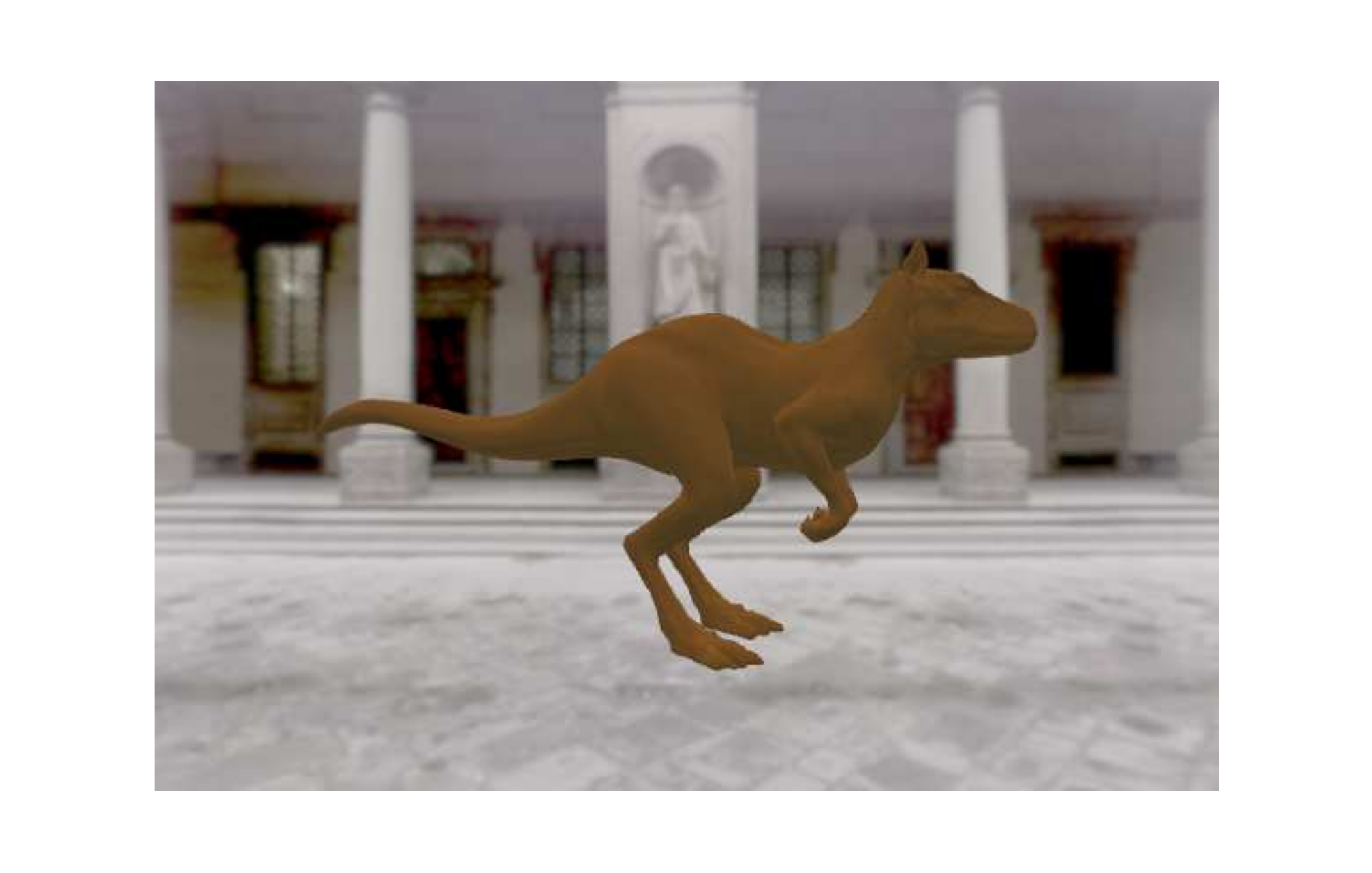}
%\vspace*{-0.4cm} 
\caption{Lambertian kangaroo lit by different environments}\label{fig:killeroo-env}
\end{figure*}

We have applied and verified our method under various lighting
scenarios for lambertian and Phong-like materials. For constant
lighting Figure \ref{fig:scenes} displays Lambertian and Phong-like
materials under such lighting. Figure \ref{fig:phong} demonstrates
the variations in the shininess of the material lit by an area light
source. Figures \ref{fig:buddha-env}, \ref{fig:dragon-env} and
\ref{fig:killeroo-env} show examples of rendering lambertian objects
in different environments. As mentioned earlier, all lambertian
objects are rendered at {\em O}(1) time complexity. For environment
lighting, one can readily see in Figures \ref{fig:buddha-env},
\ref{fig:dragon-env} and \ref{fig:killeroo-env} that our method
realistically captures the effect of colors in the environment on
the rendered objects.

\begin{figure}[h]
\begin{center}
\begin{tabular}{cc}
\epsfxsize=8cm\epsfbox{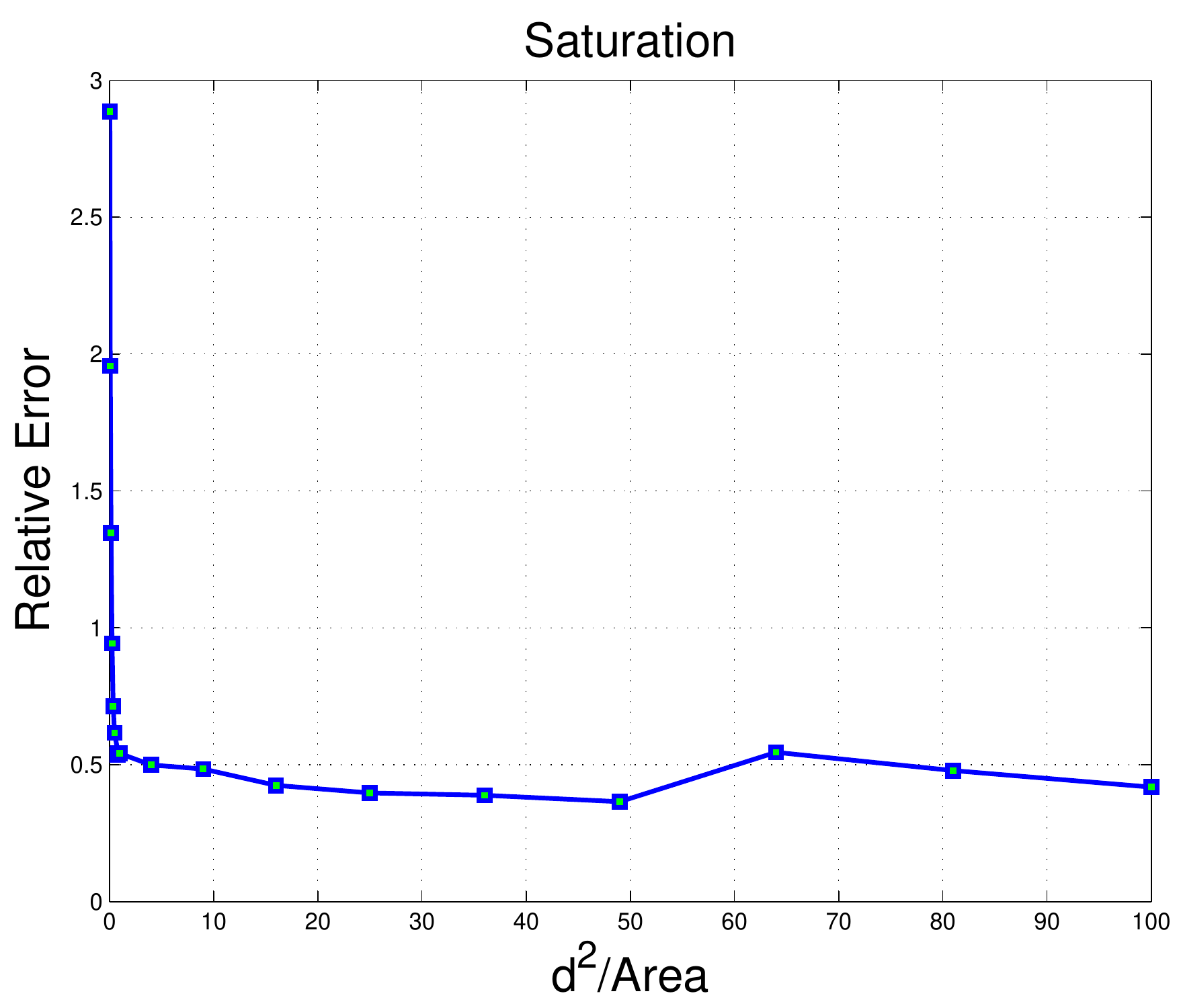} & \epsfxsize=8cm\epsfbox{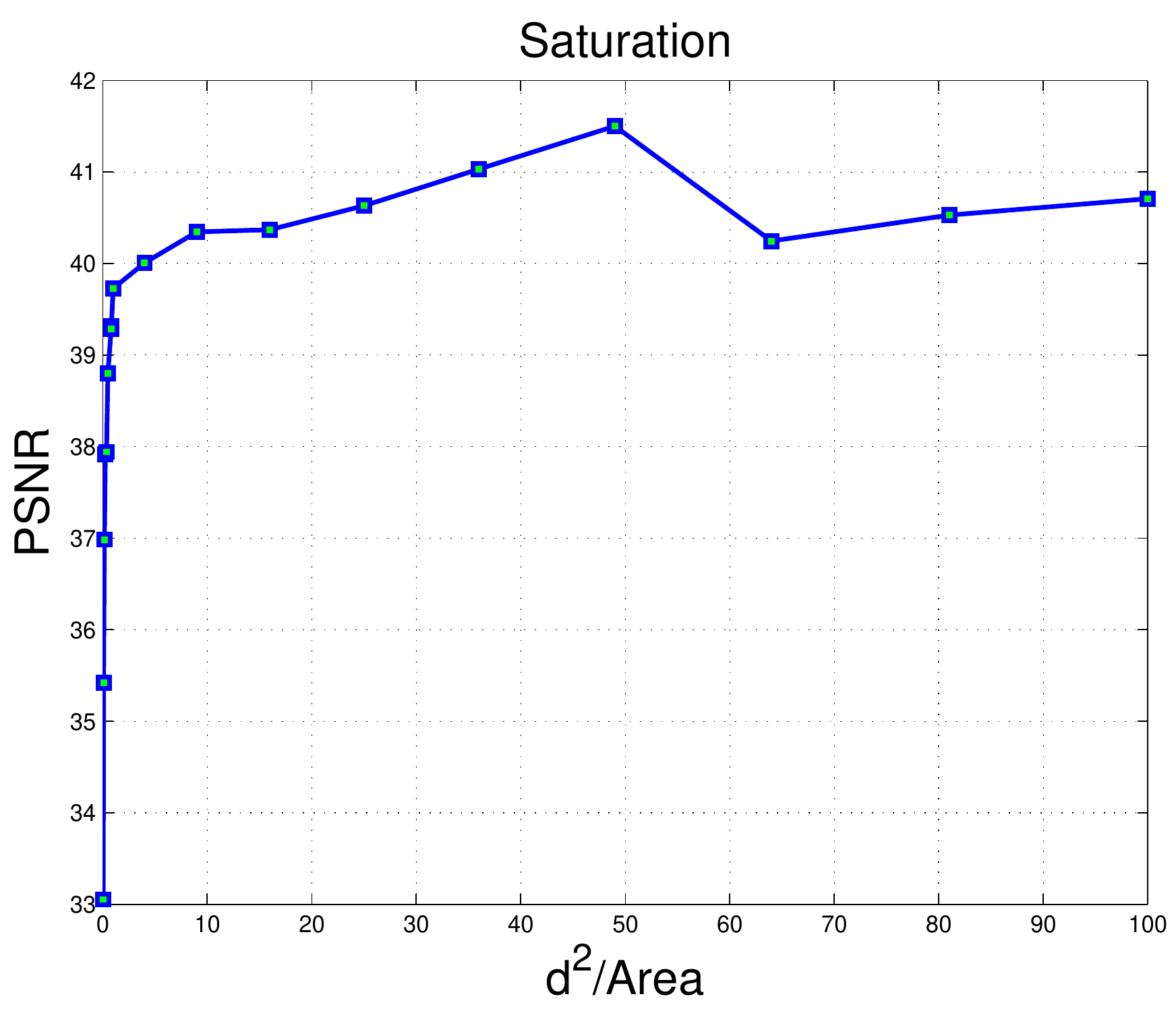} \\
(a) & (b) \\
\end{tabular}
\caption{(a) Percentage relative error of our method against Monte
Carlo, (b) Peak-Signal-to-noise error of our method against Monte
Carlo}\label{fig:PSNR}
\end{center}
\end{figure}

To evaluate the quality of our results, we performed three sets of
experiments. The first set establishes the error associated with the
approximation that led to our closed-form solution, i.e. the
assumption that the distance between the point being shaded and the
light patch $dA$ is constant and equal to the average distance. We
varied the ratio of the squared distance over the area of the light
source, $d^2/Area$, and generated ground truth images under
illumination with area light sources at finite distances, using the
Monte Carlo technique with more than 1000 samples. Then, images under
the exact same lighting conditions were generated using our
method. The images were compared with the ground truth in the HSI
domain to avoid the color channel correlations of the RGB model.
Figures \ref{fig:PSNR} - (a) and (b) show the results for the Saturation
channel.  As for the Hue and Intensity channels, we found 0\% error
indicating that for our method they are independent of the ratio
$d^2/Area$. As shown in these figures, the peak-signal-to-noise ratio
(PSNR) increases as a function of $d^2/Area$, or equivalently the
relative error reduces as this ratio increases. Note that the PSNR
increases sharply to about 40dB for $d^2/Area <1$, and remains
approximately stable thereafter. Note also that even for small values
of $d^2/Area =0.4$, the PSNR for our method has a remarkably high
value of over 33dB.

In the second set of experiments, we generated ground truth images
by rendering a lambertian sphere in several cubemaps, i.e. under
distant environment lighting. We then rendered images of the sphere
under the same lighting using our method. We tested for both SNR and
relative error against the ground-truth generated by Monte Carlo,
and averaged over a large number of environments. Results are
summarized in Table \ref{tab:error}. We observed that for lambertian
materials the lower-bound in the relative error (or the upper-bound
for SNR) is achieved with one coefficient per face in our method. In
other words, we achieve our best result with the DC value of the
cosine transform, and our method is almost invariant to the
frequency of light in the case of lambertian material.

\begin{table}[htbp]
\small
\begin{center}
\begin{tabular}{|c|c|c|}
  \hline
  RGB: Relative Error & HSI: Relative Error  & HSI: SNR (dB) \\
  \hline \hline
  R=0.0075\% & H=0.0161\% & H=42.5401 \\
  G=0.0075\% & S=0.3614\% & S=54.9891 \\
  B=0.0049\% & I=0\% & I= $\infty$ \\
\hline
\end{tabular}
\end{center}
\normalsize \caption{Errors and SNR for different color channels
averaged over several objects in different
environments.}\label{tab:error}
\end{table}

\begin{figure}[h]
\begin{center}
\begin{tabular}{cc}
\epsfxsize=8cm\epsfbox{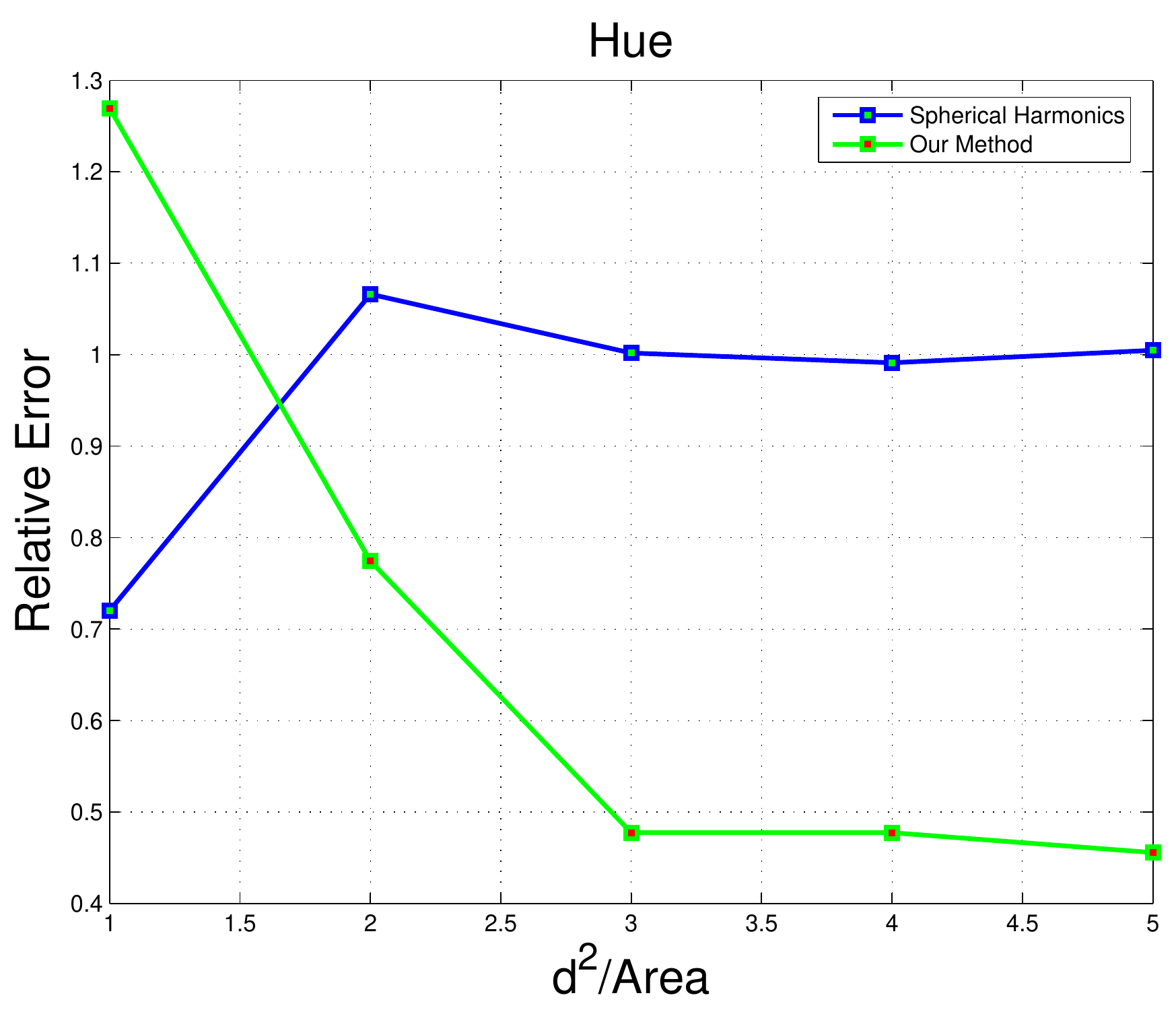} & \epsfxsize=8cm\epsfbox{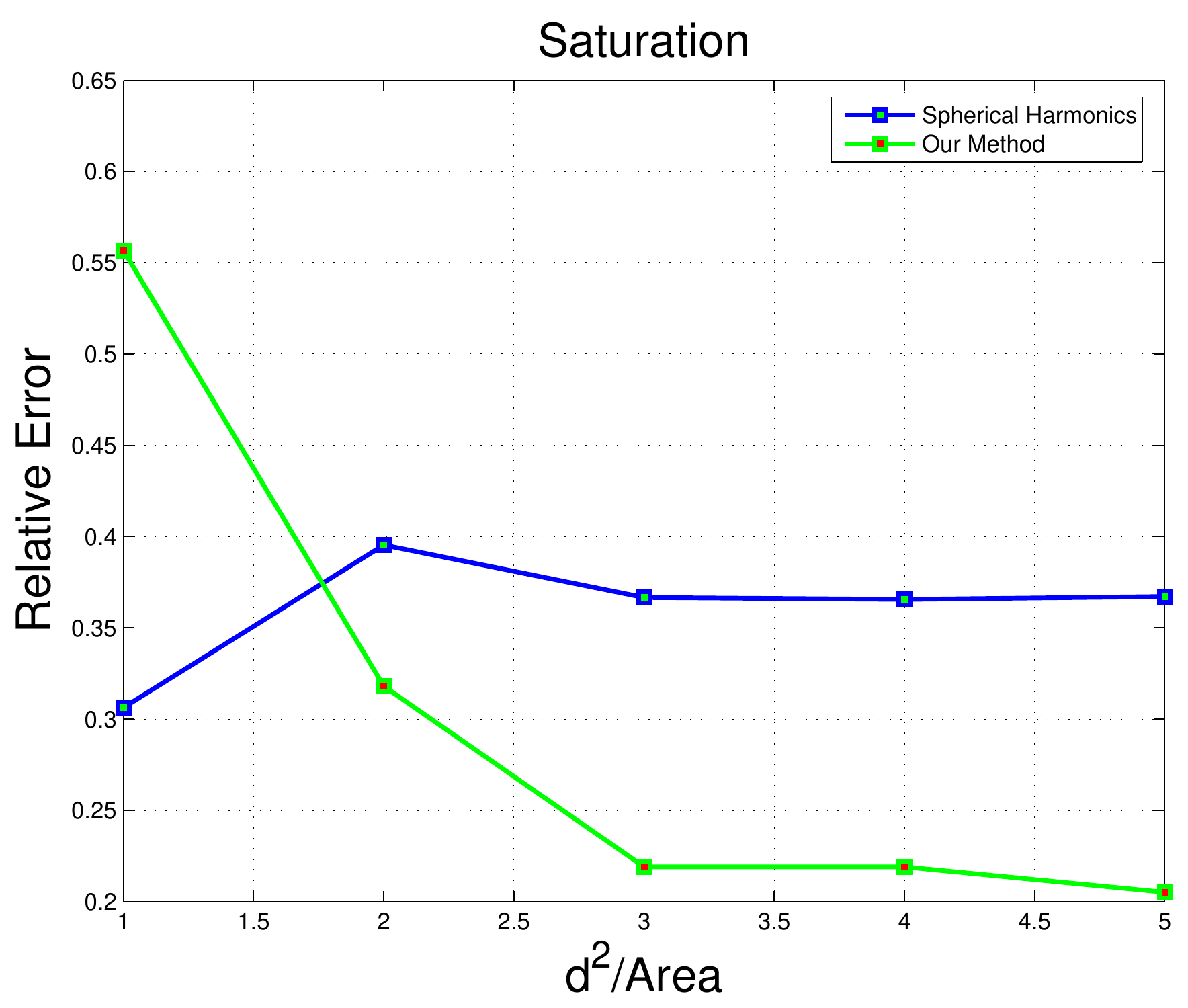} \\
(a) & (b) \\
\end{tabular}
\caption{(a) Hue Percentage relative error of Spherical Harmonics and
our method against Monte Carlo, (b) Saturation Percentage relative
error of Spherical Harmonics and our method against Monte
Carlo}\label{fig:Hrel}
\end{center}
\end{figure}

In the third set of experiments, we generated ground truth images for
multiple environments using Monte Carlo, with more than 1000
samples. We generated images under the same environments using
Spherical Harmonics and our method. The error was computed against the
ground truth provided by Monte Carlo. Figures \ref{fig:Hrel}- (a) and
(b) show the results for one of the environments used in the
experiment. By examining the figures one can tell that our method
provides more accurate results than Spherical Harmonics.

In summary, we first formulated a novel approach to provide
closed-form solutions to the light integral for lambertian and
Phong-like materials, which are lit by constant area light sources.
Spot light and directional linear light sources are then special
cases of our formulation.  Using the key observation that a cubemap
can be represented as six non-constant area light sources, we have
also shown that the same framework can be extended to provide a
closed-form solution to environment lighting. In our formulation,
the cut-off frequency of the light can be chosen arbitrarily at any
desired value in the cosine transform domain. In particular, for
lambertian material, we only require one DCT coefficient per face to
render realistic images, achieving a complexity of \emph{O}(1).
Also, we gain an order of magnitude or higher in rendering time
compared to classical environment sampling techniques such as Monte
Carlo.

In principle, models of other material types can be formulated
within our framework and solved in similar fashions. Another
direction of research could be the incorporation of general BRDFs.
One issue that needs to be considered is the rendering of shadows.
In classical estimation techniques based on sampling, shadowing is
implicitly incorporated. In our framework, due to the closed-form
nature of our solution, shadows need to be computed explicitly.
However, several methods have already been proposed in the existing
literature \cite{ChFe92,192207,Tanaka_Takahashi97} for efficient and
realistic computation of shadows that would nicely fit in our
framework.

\newpage
\clearpage
\section*{Appendix}
The following is \emph{pseudo code} for implementing the algorithm.
It is intended to provide a global understanding of our algorithm.

%\large
\begin{description}
  \item \emph{For each point $p$ to be rendered}
  \begin{description}
    \item \emph{if (lambertian)}\{
    \begin{description}
        \item \emph{Transform the end points of the light source $a_u$, $b_u$ and $c_u$
and its normal $N_{A_u}$ such that the point $p$ is at the origin
and the unit normal $\vec{N}$ at the point $p$ is along the z-axis.}
      \item \emph{Compute $l_{00}$, $l_{01}$, $l_{02}$, $l_{10}$, $l_{11}$, $l_{20}$}
      \item \emph{if($a_z\geq 0$ and $b_z<0$ and $c_z<0$ and $d_z<0$)}
         \begin{description}
          \item \emph{if(Constant)}
           \begin{description}
            \item \emph{call formula for constant\_diff\_case1.}
           \end{description}
          \item \emph{else}
           \begin{description}
            \item \emph{for each of the six faces of the cube map call formula for nonconstant\_diff\_case1.}
           \end{description}
         \end{description}
      \item \emph{else if($a_z\geq 0$ and $b_z<0$ and $c_z\geq 0$ and $d_z<0$)}
         \begin{description}
          \item \emph{if(Constant)}
           \begin{description}
            \item \emph{call formula for constant\_diff\_case2.}
           \end{description}
          \item \emph{else}
           \begin{description}
            \item \emph{for each of the six faces of the cube map call formula for nonconstant\_diff\_case2.}
           \end{description}
         \end{description}
      \item \emph{else if(\ldots)}
      \item :
      \item \emph{else if(($a_z<0$ and $b_z<0$ and $c_z<0$ and $d_z<0$)}
        \begin{description}
         \item \emph{return 0}
        \end{description}
    \end{description}
    \item \}
    \item \emph{else if(specular)}\{
     \begin{description}
      \item \emph{Transform the end points of the light source $a_u$, $b_u$ and $c_u$
and its normal $N_{A_u}$ such that the point $p$ is at the origin
and the reflection vector $\vec{R}$ at the point $p$ is along the
z-axis.}
      \item \emph{Compute $l_{00}$, $l_{01}$, $l_{02}$, $l_{10}$, $l_{11}$, $l_{20}$}
        \item \emph{if($a_z\geq 0$ and $b_z<0$ and $c_z<0$ and $d_z<0$)}
         \begin{description}
          \item \emph{call formula for constant\_spec\_case1.}
         \end{description}
       \item \emph{if($a_z\geq 0$ and $b_z<0$ and $c_z\geq 0$ and $d_z<0$)}
        \begin{description}
         \item \emph{call formula for constant\_spec\_case2.}
        \end{description}
       \item \emph{else if(\ldots)}
       \item :
       \item \emph{else if($a_z<0$ and $b_z<0$ and $c_z<0$ and $d_z<0$)}
        \begin{description}
         \item \emph{return 0}
        \end{description}
     \end{description}
    \item \}
  \end{description}
\end{description}
\small

\clearpage
\bibliographystyle{plain}
\bibliography{foroosh,bibliography}

\end{document}